\documentclass[10pt,twocolumn,letterpaper]{article}

\usepackage{cvpr}              
\usepackage{multirow}
\usepackage{arydshln}
\definecolor{cvprblue}{rgb}{0.21,0.49,0.74}
\usepackage[pagebackref,breaklinks,colorlinks,allcolors=cvprblue] {hyperref}     
\newcommand{\pparagraph}[1]    {\smallskip\noindent\textbf{#1}}
\def\confName{CVPR}  
\def\confYear{2027}

\title{GeomSat: Learning Geometric-Aware Representations for Very High Resolution Satellite Imagery}

\author{%
  Xavier Bou\textsuperscript{1} \qquad
  Gabriele Facciolo\textsuperscript{2,3} \qquad
  Thibaud Ehret\textsuperscript{1}\\[0.4em]
  {\small \textsuperscript{1}AMIAD, France}\\
  {\small \textsuperscript{2}Université Paris-Saclay, CNRS,
   ENS Paris-Saclay, Centre Borelli, France}\\
   {\small \textsuperscript{3}Institut Universitaire de France, France}
   \\[0.3em]
  {\footnotesize\ttfamily
   \url{https://xavibou.github.io/geomsat_pages}}%
}

\makeatletter
\let\originalmaketitle\@maketitle
\renewcommand{\@maketitle}{%
    \originalmaketitle
    \noindent
    \begin{minipage}{\textwidth}
        \centering
       \includegraphics[
        width=\linewidth,
        trim=0bp 0bp 0bp 0bp,
        clip
    ]{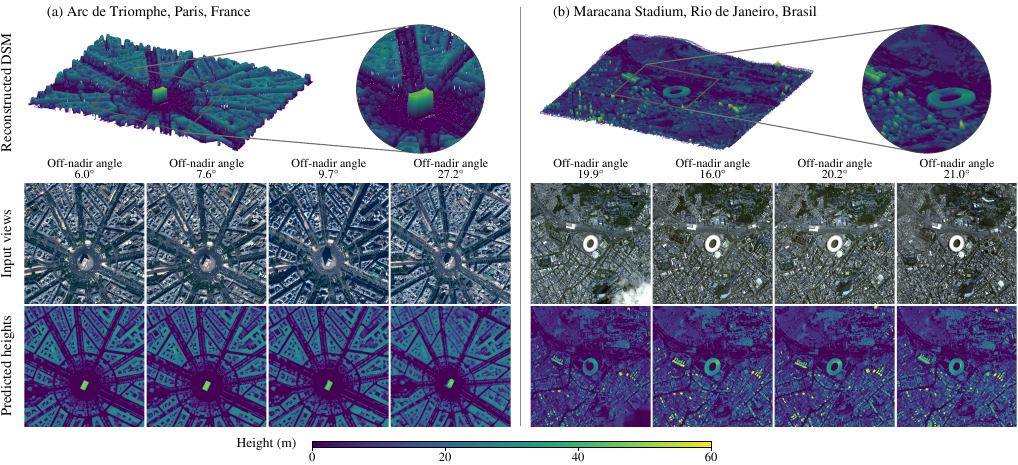}
    
    \vspace{-.5em}
    
        \captionof{figure}{\textbf{From satellite imagery to 3D surface reconstruction}.
        Qualitative results for the Arc de Triomphe in Paris (left) and Maracana Stadium in Rio de Janeiro (right) for monocular height estimation.
        Our model predicts height above ground for each available view. These predictions are then fused across acquisitions via their adjusted RPC models, producing a reconstructed Digital Surface Model (DSM).}
        \label{fig:teaser}
    \end{minipage}
    \par\vspace{0.5em}
}
\makeatother

\begin{document}
\maketitle
\begin{abstract}
Very-high-resolution (VHR) satellite imagery enables applications such as disaster management or environmental monitoring. Using these observations effectively requires recovering the geometry of the scene despite variations in viewpoint, illumination, and atmospheric conditions. Existing satellite reconstruction methods typically require multiple images and per-scene optimization, while general-purpose geometric models developed for natural images transfer poorly to satellite data. We introduce GeomSat, an automated, scalable pipeline for geometric learning in this domain, consisting of data generation, self-supervised pretraining, and fine-tuning on two geometric tasks: monocular height estimation and dense correspondence. Our model enables digital surface model reconstruction by combining height predictions across views. Experiments show strong pre-trained geometric representations and improvements over state-of-the-art baselines in both fine-tuned tasks. 
\end{abstract}    
\section{Introduction}
\label{sec:intro}
\vspace{-0.5em}
Very-high-resolution (VHR) satellite imagery enables the observation of the Earth's surface at sub-meter resolution, revealing individual buildings, roads, and vegetation over the globe. 
This level of detail supports a wide range of applications,
such as urban mapping \cite{gibril2024transformer},
environmental monitoring \cite{tucker2023carbon},
infrastructure assessment \cite{workman2023prediction},
and disaster response \cite{hafner2025disasteradaptivenet}. For many of these, understanding the observed environment requires more than identifying elements in the image: their height, spatial arrangement, and three-dimensional structure are also essential.

However, reasoning about scene structure from VHR satellite imagery presents particular challenges. Observations of the same location are often acquired from different points of view. Thus, elevated structures appear to be displaced differently with respect to their ground footprints, different surfaces become visible, and occlusions vary across views. Images covering the same area therefore do not spatially align.
On top of geometric differences, acquisitions may differ in illumination, shadows, vegetation, cloud cover, and semantic content. Thus, preparing VHR imagery for geometric analysis involves substantial processing, with dedicated pipelines to handle panchromatic and multispectral products, refine satellite camera models~\cite{mari2019bundle, mari2021generic}, and relate observations across views~\cite{masquil2026diachronic, diaz2026seasonstereo}. Together, these factors make it difficult to match and combine information between observations of the same area.

A natural way to relate these observations is via a three-dimensional representation of the observed surface, providing a common geometric reference in which image locations can be associated with physical scene points. 
Stereo and multi-view stereo methods recover surface geometry by matching corresponding regions across views~\cite{bosch2019semantic,gao2021satmvs}. However, illumination, seasonal, and scene variations between acquisitions can reduce the reliability of these matches~\cite{bosch2019semantic,mari2022satnerf}.
Satellite neural rendering methods 
address illumination changes and transient effects~\cite{derksen2021snerf,mari2022satnerf,mari2023eonerf}, with later work  exploring shadow supervision~\cite{masquil2025seo} and learned diachronic matching~\cite{masquil2026diachronic,diaz2026seasonstereo}. Nevertheless, some approaches require scene-specific optimization, while diachronic stereo~\cite{masquil2026diachronic} focuses on pairwise disparity estimation rather than learning transferable geometric representations across scenes.

Large-scale geometric learning offers a way to acquire such representations. Multi-view photo collections provide supervision for monocular depth learning, as demonstrated by MegaDepth~\cite{li2018megadepth}, while Depth Anything~\cite{yang2024depthanything,yang2024depthanythingv2} and Metric3D v2~\cite{yang2024depthanything} learn transferable priors for single-image geometry. More recent approaches, including DUSt3R~\cite{wang2024dust3r}, VGGT~\cite{wang2025vggt}, VGGT-$\Omega$~\cite{wang2026vggtomega}, Depth Anything 3~\cite{lin2025depthanything3} and D4RT~\cite{Zhang_2026_CVPR}, extend geometric prediction to multiple observations. These advances primarily stem from ground-level photographs, synthetic scenes, and video, failing to maintain their performance on remote sensing data. This motivates learning representations that consider camera geometry, spatial sampling, and temporal appearance variations of VHR satellite imagery.

In this work, we develop a recipe for geometric learning in VHR satellite imagery. We first construct a geographically diverse multi-view dataset from Pleiades~\cite{pleiades_product_guide} acquisitions distributed across the globe. Starting from raw products, we derive a scalable and automatic pipeline generating 1.8 million multi-view observations, comprising around 9 million individual images. We further complement these data with metric supervision that uses available LiDAR products on a subset of covered areas. We project point clouds into their corresponding satellite acquisitions and generate dense geometric annotations in each image's viewing geometry. These annotations include metric above-ground height maps (or normalized digital surface modes), surface normal fields, and above ground-displacement maps. This yields a densely supervised subset of the broader, geographically diverse collection of multi-view observations. We name the generated corpus of geometric satellite data as the GeomSat dataset.

We leverage these data for geometric tasks: we first pre-train an image encoder via cross-view completion~\cite{weinzaepfel2023croco}, learning domain-specific features that relate scene appearance across viewpoints. Then, using the LiDAR-derived labels, we fine-tune these representations on two geometric tasks, namely monocular height estimation and dense correspondence prediction. Our experiments show that our data and training methodology yield robust geometric models that generalize on other  geographical areas.
Our contributions can be summarized as:
\begin{itemize}
    \item We introduce GeomSat: an automatic and scalable pipeline to process raw satellite products and turn them into multi-view geometric data for self-supervised learning. It leverages LiDAR coverage to generate dense geometric labels for these observations. 
    \item We pre-train a GeomSat encoder on the multi--view samples, learning geometric satellite-specific representations. We then fine-tune them on two geometric tasks: monocular height estimation and dense correspondence.
    \item Via several experiments we show the potential of GeomSat at scale: transferable geometric features, and fine-tuned models achieving competitive results with a modest parameter count (see Fig.~\ref{fig:teaser}).
\end{itemize}

\vspace{-0.5em}
\section{Related Works}
\vspace{-0.5em}

\pparagraph{Learning visual geometry.}
Large-scale image collections have enabled geometric representations that transfer across scenes. MegaDepth~\cite{li2018megadepth} demonstrated monocular depth learning from internet-photo reconstructions, while Depth Anything V2~\cite{yang2024depthanythingv2}, Metric3D v2~\cite{hu2024metric3dv2}, and MoGe~\cite{wang2025moge} learn increasingly general single-image geometric priors. Beyond monocular prediction, CroCo and CroCo v2~\cite{weinzaepfel2022croco,weinzaepfel2023croco} learn correspondence-sensitive features through cross-view completion, while Alligat0R~\cite{loiseau2025alligat0r} replaces reconstruction with covisibility segmentation to better handle image pairs with limited overlap. Building on cross-view representations, DUSt3R~\cite{wang2024dust3r} and MASt3R~\cite{leroy2024mast3r} extend geometric learning toward point-map regression and 3D-aware matching, ultimately leading to  VGGT~\cite{wang2025vggt}, VGGT-$\Omega$~\cite{wang2026vggtomega}, Depth Anything~3~\cite{lin2025depthanything3}, and D4RT~\cite{Zhang_2026_CVPR} which enable feed-forward geometric inference from multiple observations. These models are mostly developed on perspective imagery, and their predicted depth is not directly equivalent to above-ground height in off-nadir satellite acquisitions. We instead learn from real multi-view satellite observations and explicitly supervise metric height in the original image geometry.

\pparagraph{Multi-view satellite reconstruction.}
Satellite photogrammetry recovers surface geometry by matching observations across viewpoints, often after refining rational polynomial coefficient (RPC) camera models~\cite{mari2019bundle,mari2021generic}. US3D~\cite{bosch2019semantic} and SatMVS~\cite{gao2021satmvs} demonstrate semantic and learning-based stereo reconstruction with satellite imagery; the S-EO dataset~\cite{masquil2025seo} further combines multi-date imagery, LiDAR surface models, and refined RPCs. Satellite neural-rendering methods, including S-NeRF, Sat-NeRF, EO-NeRF, and EO-GS~\cite{derksen2021snerf,mari2022satnerf,mari2023eonerf,aira2025gaussian}, model view-dependent appearance and illumination but typically optimize a representation for each scene. More recent work moves toward transferable reconstruction: SkySplat~\cite{huang2026skysplat} learns generalizable, RPC-aware reconstruction from sparse multi-temporal images, and EO-VGGT~\cite{luo2026eovggt} adapts a geometric foundation model through orbital ray conditioning. Unlike approaches that reconstruct a scene from multiple inputs at test time, our multi-view observations provide pretraining supervision for geometric prediction from a single image.

\pparagraph{Monocular geometry in remote sensing.}
Estimating absolute height from one overhead image is intrinsically ambiguous and has motivated specialized architectures and benchmarks. HTC-DC Net~\cite{chen2023htcdc} addresses long-tailed height distributions, THE Benchmark~\cite{xiong2023the} studies cross-dataset transfer, and SynRS3D/RS3DAda~\cite{song2024synrs3d} use synthetic scenes and domain adaptation for monocular 3D understanding. Others use a geometry priors, including RDAH-Net~\cite{jiang2026rdah} and application-specific canopy-height models such as CHMv2~\cite{brandt2026chmv2}. Importantly, view-dependent displacement has also been investigated: Christie \emph{et al.}~\cite{christie2020geocentric} jointly predict height and dense parallax-correcting flow for oblique satellite imagery, while monocular building-reconstruction methods~\cite{li2021monocularbuilding,li2024multilevelbuilding} estimate roof-to-footprint offsets. 

\pparagraph{Geospatial pretraining and geometry.}
Earth-observation pretraining has addressed the domain gap through temporal and multispectral masked modeling in SatMAE~\cite{cong2022satmae}, scale-aware reconstruction in Scale-MAE~\cite{reed2023scalemae}, and large multi-modal pretraining in SkySense~\cite{guo2024skysense}. These objectives improve geospatial representations but do not directly enforce cross-view 3D consistency. Geometry-oriented resources include the LiDAR-derived image--height pairs of GlobalBuildingAtlas~\cite{zhu2025globalbuildingatlas} and HiRes-FusedMIM~\cite{mutreja2026hires}, which studies large-scale RGB--DSM pretraining. Our data construction is complementary: it provides approximately 1.83 million co-located groups comprising 8.96 million real Pleiades views from geographically distributed sites. 
\begin{figure}[t]
    \centering
    \includegraphics[width=\columnwidth]{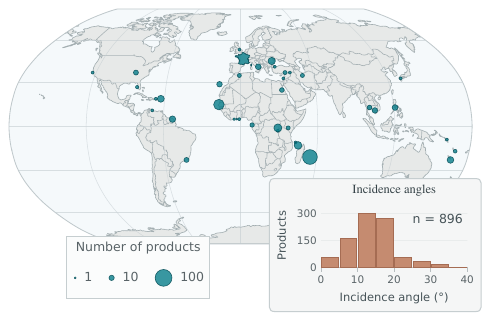}
    \caption{
        \textbf{Geographic and viewing-angle distribution of the dataset.}
        The world map shows the spatial distribution of Pleiades acquisitions. The incidence-angle histogram summarizes the diversity of viewing geometries represented in the dataset.
        \vspace{-2em}
    }
    \label{fig:dataset_distribution}
\end{figure}
\vspace{-0.5em}
\section{The GeomSat Training Dataset}
\vspace{-0.5em}
Our training dataset is built from raw VHR Pleiades~\cite{pleiades_product_guide} products. 
We identify urban areas with a number of available acquisitions and retain  sites containing at least three products with overlapping footprints. We pansharpen them and, for each site, bundle adjustment is jointly performed over overlapping observations to refine their RPC camera models and improve cross-view geometric consistency, using the approach of Mari \textit{et al.}~\cite{mari2026robust}.
We select reference acquisitions that maximize overlap with the remaining observations of each site and extract \(512\times512\) crops using a sliding window with 128-pixel overlap. For each reference crop, we project its geographic location onto the remaining acquisitions using the refined RPC models, yielding sets of crops that observe the same area under different viewing geometries. Crops dominated by water are detected using the Normalized Difference Water Index (NDWI) and discarded.

\pparagraph{Statistics.}
The resulting dataset contains 925 unique Pleiades images grouped into 152 sites covering approximately \(9{,}573\,\mathrm{km}^2\) of non-overlapping area.  Each site is made up of observations from multiple dates, some spanning different seasons.
After crop extraction and filtering, the dataset comprises \(8{,}961{,}006\) images and \(1{,}833{,}753\) samples.
We partition the dataset into geographically disjoint training, validation, and test sets using \(80\%\), \(10\%\), and \(10\%\) splits, respectively, yielding 107 training sites, 22 validation sites, and 23 test sites. Figure~\ref{fig:dataset_distribution} shows the geographic distribution of the selected acquisitions together with their incidence-angle distribution.

\pparagraph{Geometric annotations.}
We generate dense metric supervision for the regions where airborne LiDAR data is available such as, for example, LiDAR HD~\cite{ign_lidarhd} data over France and USGS 3D Elevation Program data~\cite{usgs_3d} over the USA. We first project the LiDAR measurements into each Pleiades acquisition using its refined RPC model, and further correct the residual image--LiDAR misalignment through feature matching and homography estimation using XoFTR~\cite{xoftr}. The projected points are rendered onto the image grid, retaining the visible surface where multiple points overlap. Gaps along building facades are completed using locally supported planar surfaces, while regions without sufficient LiDAR support remain masked. We use height above ground, i.e. the difference between Digital Surface Model (DSM) and Digital Terrain Model (DTM), also known as normalized Digital Surface Model (nDSM).

In off-nadir acquisitions, elevated points are displaced from their ground footprint by an amount that grows with height and depends on the viewing geometry, similar to stereo disparity. 
We therefore define a \emph{above-ground view} as a canonical reference and encode it as a \emph{above-ground displacement field}: a per-pixel 2D vector from the observed image position to where the same 3D point would appear if the point were on the ground.
This gives a single-image, view-dependent geometric cue whose supervision is analogous to cross-view displacement, and which complements the metric height map. Height captures the vertical structure, while the displacement field captures how that structure is projected under the current acquisition geometry. We generate this label for every image with LiDAR supervision and additionally derive surface normal fields from the same LiDAR-derived labels, following recent work~\cite{wang2025moge}.

\vspace{-0.5em}
\section{Geometry-aware model for remote sensing}
\label{sec:method}
\vspace{-0.25em}
\subsection{Learning generic geometry aware features}
\vspace{-0.25em}
\pparagraph{Overview.}
We aim to leverage the generated multi-view observations and labels for geometric tasks. We follow a two-stage approach. First, we pre-train a visual encoder via self-supervised cross-view reconstruction. Then, we fine-tune the learned features using the geometric labels for two geometric tasks: monocular height estimation and dense correspondence prediction. We expand on each below.

\pparagraph{Pre-training.}
We adopt the CroCo v2 cross-view completion approach~\cite{weinzaepfel2022croco,weinzaepfel2023croco}, comprising a Vision Transformer encoder and a cross-view decoder. Given a multi-view sample, we randomly select two observations $I_q$ and $I_r$, which correspond to the same geographical area and were acquired from different viewpoints. We randomly mask a subset $\mathcal{M}$ of patches in the query image $I_q$. The encoder independently extracts features from visible query patches and the unmasked reference image $I_r$, and the decoder uses both representations to reconstruct the masked query content. The pretraining objective is
\begin{equation}
\mathcal{L}_{\mathrm{pre}} =
\frac{1}{|\mathcal{M}|}
\sum_{p\in\mathcal{M}}
\ell_{\mathrm{rec}}\!\left(\hat{\mathbf{x}}_p,\mathbf{x}_p\right),
\label{eq:pretrain}
\end{equation}
where $\mathbf{x}_p$ and $\hat{\mathbf{x}}_p$ are the target and reconstructed content of the masked patch $p$, respectively. The reconstruction loss is $\ell_{\mathrm{rec}}(\hat{\mathbf{x}}_p,\mathbf{x}_p) = \frac{1}{D}\|\hat{\mathbf{x}}_p-\mathbf{x}_p\|_2^2$, where $D$ is the number of pixel values in each patch and the target values are normalized using the patch's mean and standard deviation. This step aims to learn visual features from our geographically diverse dataset without requiring task-specific annotations.
\vspace{-0.25em}

\subsection{Application-specific heads}
\vspace{-0.25em}
\pparagraph{Monocular height estimation.} 
Using the generated geometric labels, we fine-tune the pre-trained encoder for monocular height estimation. 
We use a DPT-style head~\cite{ranftl2021vision} that combines features from multiple encoder layers through coarse-to-fine fusion. Acquisition metadata, including viewing geometry, ground sampling distance (GSD), and solar azimuth and elevation, provide meaningful priors for estimating heights. Viewing geometry determines the perceived displacement of elevated structures, whereas illumination determines shadow geometry, directly relating their area to the height of its structures. We therefore incorporate this information into the head through feature-wise linear modulation (FiLM)~\cite{perez2018film}, which we show to considerably improve performance. Details of the head architecture and exact metadata values are provided in the supplementary material.


Given a single image $I$ and its acquisition metadata $\mathbf{m}$, the model predicts an above-ground height map $\hat H$, rather than an absolute elevation, since the elevation of the terrain cannot be reliably inferred from a single observation. In addition, we experiment with auxiliary supervision using the generated geometric labels, jointly predicting a two-channel nadir-displacement field $\hat{\mathbf D}$, a three-channel surface-normal field $\hat{\mathbf N}$, or both. 
Near-nadir observations provide limited visual cues about vertical structures, as the sides of elevated objects are essentially invisible. To complement these cues, above-ground displacement supervision encourages the model to learn the relationship between object height and acquisition geometry, determining the apparent displacement of elevated structures relative to their nadir positions. Similarly, surface normals encourage the model to capture local surface orientation and geometric variations, providing complementary cues about the shape and structure of objects. We supervise height and above-ground displacement with $\ell_1$ losses, averaging over the two displacement components, and surface normals with cosine distance after unit normalization, following Eigen and Fergus~\cite{eigen2015predicting}. Each task loss is computed over its valid annotated pixels:
\begin{equation}
    \mathcal L_t =  
    \frac{\sum_{\mathbf p\in\mathcal V_t}
    w(\mathbf p)\,\ell_t(\mathbf p)}
    {\sum_{\mathbf p\in\mathcal V_t}w(\mathbf p)},
    \qquad t\in\{H,D,N\},
\end{equation}
where $\mathcal V_t$ denotes the set of valid pixels for task $t$, $\ell_t(\mathbf p)$ is the corresponding loss per-pixel, and $w(\mathbf p)$ is an inverse-frequency weight determined by the ground-truth height bin of pixel $\mathbf p$. The weights are computed from the distribution of the training-set height to compensate for the predominance of low-height pixels. We fine-tune both pre-trained encoder and head. The overall objective combines height estimation with the enabled
auxiliary tasks:
\begin{equation}
    \mathcal L =
    \mathcal L_H
    + \lambda_D\,\mathcal L_D
    + \lambda_N\,\mathcal L_N,
    \label{eq:height_finetune}
\end{equation}
where $\lambda_D$ and $\lambda_N$ control the weight of
displacement and normal supervision, respectively.

\pparagraph{Dense correspondence prediction.}
We also fine-tune the pre-trained encoder to establish correspondences
between satellite images acquired from different viewpoints, similarly to methods like XoFTR~\cite{xoftr} or SuperGlue~\cite{sarlin2020superpoint}. We extract
features from multiple encoder layers and project them into a
multi-scale feature pyramid. As in height estimation, we condition
these features on each image's acquisition metadata through FiLM layers.
We adopt the coarse-to-fine matching modules of XoFTR~\cite{xoftr},
which use self- and cross-attention to establish coarse matches and
progressively refine their locations to subpixel precision. The
encoder and matching modules are jointly fine-tuned.

Given two images $I_q$ and $I_r$ and their acquisition metadata,
the model predicts a set of correspondences
$\{(\hat{\mathbf p}_i,\hat{\mathbf q}_i)\}$.
We supervise these predictions using the generated geometric
correspondences, restricting the supervision to locations with known
geometric validity. Coarse and fine matching scores are supervised
with focal losses, while a squared coordinate error supervises
subpixel refinement:
\begin{equation}
    \mathcal L_{\mathrm{match}} =
    \lambda_c\,\mathcal L_{\mathrm{coarse}}
    + \lambda_f\,\mathcal L_{\mathrm{fine}}
    + \lambda_s\,\mathcal L_{\mathrm{subpixel}},
    \label{eq:matching_finetune}
\end{equation}
where $\lambda_c$, $\lambda_f$, and $\lambda_s$ balance the three
terms. The subpixel loss is applied to geometrically valid matches
whose predicted endpoints lie within a local neighborhood of their
targets. At inference, correspondence prediction requires only the
two images and their acquisition metadata.
\vspace{-0.5em}

\section{Experiments}
\vspace{-0.25em}
\pparagraph{Implementation details.}
We pre-train a ViT-Base encoder with 12 transformer blocks, 768-dimensional embeddings, and 12 attention heads, paired with a cross-view decoder with 8 blocks, 512-dimensional embeddings, and 16 attention heads. Following CroCo v2~\cite{weinzaepfel2023croco}, both the encoder and decoder use two-dimensional rotary positional embeddings. We randomly mask 90\% of the query patches on the images of size $512\times512$. 
We use AdamW optimizer with $\beta=(0.9,0.95)$, weight decay $0.05$, and a batch size of 56 image pairs per GPU. The learning rate increases linearly over 40 warm-up epochs to $5.25\times10^{-4}$, followed by cosine decay with an 800-epoch horizon. Pretraining ran for 400 epochs on 16 B200 NVIDIA GPUs, taking approximately 5.7 days in total. 
For height-estimation, we attach a randomly initialized DPT head to the pre-trained encoder and optimize both jointly using AdamW with $\beta=(0.9,0.95)$ and weight decay $0.05$. The encoder and head use peak learning rates of $10^{-5}$ and $10^{-4}$, respectively, with five warm-up epochs followed by cosine decay over a 120-epoch schedule and a batch size of 64. We apply random square crops covering 80--100\% of the image area, resize them to $512\times512$, and add random rotations from $[-180^\circ,180^\circ]$. Color augmentation independently perturbs the gain and contrast of each channel within $[0.9,1.1]$ and $[0.95,1.05]$, respectively. We optimize an $\ell_1$ height loss together with an auxiliary surface-normal cosine loss weighted by $0.1$. Both losses use inverse-frequency weights computed from training-set height bins, with seven 5\,m intervals spanning 0--35\,m and a final bin above 35\,m. We evaluate these architectural and training choices via ablations in the experiments and provide further implementation details in the supplementary. For dense correspondence, we follow XoFTR~\cite{xoftr} and set the loss weights to $\lambda_c=0.5$, $\lambda_f=0.3$, and $\lambda_s=10^{4}$.

\pparagraph{Evaluation protocol.}
For self-supervised pre-training, we combine the training and validation splits while keeping the test sites held out. We evaluate geometric tasks on the LiDAR-annotated subset: models are fine-tuned on 150k training images with geometric annotations and evaluated on 506,015 images from the held-out test sites. We also assess cross-dataset generalization on US3D/DFC19~\cite{bosch2019semantic}, using WorldView-3 imagery to evaluate transfer across sensors and spatial resolutions without fine-tuning on the target dataset. We compare our approach with relevant state-of-the-art methods for monocular height estimation and dense correspondence prediction.

\begin{table*}[t]
\centering
\begin{minipage}[t]{0.695\textwidth}
\vspace{0pt}\centering
{%
\caption{\textbf{Monocular height estimation.} MAE (m, $\downarrow$) on GeomSat and US3D/DFC19 for all pixels and GT heights $\geq20$\,m, split by vegetation. Aligned predictions use a per-image signed scale and offset fitted to GT. Best and second-best results are \textbf{bold} and \underline{underlined}. $^\dagger$CHMv2 specializes in canopy height; RS3DAda was trained on crops of the same US3D products. Compared methods~\cite{yang2024depthanythingv2,wang2025vggt,
wang2025moge,wang2026vggtomega,hu2024metric3dv2,
lin2025depthanything3,song2024synrs3d,brandt2026chmv2}. RMSE is in the supplementary.\vspace{-.5em}}
\label{tab:height-benchmark}
\renewcommand{\arraystretch}{1.14}
\setlength{\tabcolsep}{1.6pt}
\fontsize{4.5}{4.5}\selectfont
\resizebox{0.97\linewidth}{!}{%
\begin{tabular}{@{}l*{6}{r}@{\hspace{5pt}}*{6}{r}@{}}
\toprule
& \multicolumn{6}{c}{\textbf{GeomSat (Pl\'eiades)}}
& \multicolumn{6}{c}{\textbf{US3D / DFC19}} \\
\cmidrule(lr){2-7}\cmidrule(l){8-13}
& \multicolumn{2}{c}{Full} & \multicolumn{2}{c}{No veg.} & \multicolumn{2}{c}{Vegetation}
& \multicolumn{2}{c}{Full} & \multicolumn{2}{c}{No veg.} & \multicolumn{2}{c}{Vegetation} \\
\cmidrule(lr){2-3}\cmidrule(lr){4-5}\cmidrule(lr){6-7}
\cmidrule(lr){8-9}\cmidrule(lr){10-11}\cmidrule(l){12-13}
Model & All & $\geq20$ & All & $\geq20$ & All & $\geq20$
      & All & $\geq20$ & All & $\geq20$ & All & $\geq20$ \\
\midrule
\multicolumn{13}{@{}l}{\textit{Aligned to GT per image}} \\
DAv2-B & 1.81 & 10.44 & 1.23 & 15.57 & 2.83 & 10.28 & 2.20 & \textbf{10.05} & 1.83 & 9.86 & 4.19 & 10.79 \\
VGGT & 1.85 & 10.63 & 1.29 & 16.09 & 2.84 & 10.46 & 2.37 & \underline{10.17} & 2.02 & \underline{9.82} & 4.24 & 11.54 \\
MoGe-3-L & 1.86 & 10.91 & 1.29 & 16.38 & 2.86 & 10.74 & 2.56 & 11.23 & 2.21 & 10.98 & 4.43 & 12.21 \\
VGGT-$\Omega$ & 1.88 & 10.94 & 1.31 & 15.74 & 2.89 & 10.79 & 2.52 & 10.24 & 2.16 & \textbf{9.76} & 4.45 & 12.11 \\
Metric3D-v2-L & 1.97 & 12.09 & 1.39 & 19.96 & 3.01 & 11.85 & 3.29 & 19.48 & 2.96 & 20.42 & 5.05 & 15.82 \\
DA3-B & 1.98 & 12.09 & 1.40 & 20.01 & 3.01 & 11.84 & 3.19 & 17.37 & 2.85 & 17.78 & 5.03 & 15.79 \\
RS3DAda (al.) & 1.49 & 9.14 & 0.92 & \underline{8.61} & 2.50 & 9.15 & \multicolumn{6}{c}{---} \\
CHMv2$^\dagger$ (al.) & 1.21 & \underline{6.20} & 0.89 & 21.06 & \underline{1.77} & \underline{5.73} & 2.85 & 19.94 & 2.76 & 22.78 & \textbf{3.32} & 8.93 \\
\textbf{Ours} (al.) & \multicolumn{6}{c}{---} & \textbf{1.98} & 11.37 & \underline{1.72} & 12.10 & \underline{3.39} & 8.57 \\
\addlinespace[3pt]
\multicolumn{13}{@{}l}{\textit{Metric predictions}} \\
CHMv2$^\dagger$ & \underline{1.18} & 6.84 & 0.64 & 24.49 & 2.13 & 6.29 & 3.17 & 29.03 & 2.48 & 34.79 & 6.88 & \textbf{6.78} \\
RS3DAda & 1.75 & 19.01 & \underline{0.58} & 11.46 & 3.81 & 19.25 & \underline{2.00} & 14.47 & \textbf{1.54} & 14.27 & 4.49 & 15.23 \\
\textbf{Ours} & \textbf{0.91} & \textbf{2.98} & \textbf{0.48} & \textbf{3.14} & \textbf{1.67} & \textbf{2.97} & 3.41 & 13.31 & 2.69 & 14.73 & 7.27 & \underline{7.81} \\
\bottomrule
\end{tabular}%
}
}
\vspace{-0.4em}
\end{minipage}\hfill%
\begin{minipage}[t]{0.285\textwidth}
\vspace{0pt}\centering
{%
\caption{\textbf{Height ordering on US3D.} Mean per-patch Spearman $\rho$ and Kendall $\tau_b$ ($\uparrow$) on valid pixels (All) and GT heights $\geq20$\,m. Undefined correlations are excluded. Best and second-best are \textbf{bold} and \underline{underlined}.}
\label{tab:us3d-height-ordering}
\renewcommand{\arraystretch}{1.14}
\setlength{\tabcolsep}{1.5pt}
\fontsize{7}{8.4}\selectfont
\resizebox{\linewidth}{!}{%
\begin{tabular}{@{}lrrrr@{}}
\toprule
& \multicolumn{2}{c}{All} & \multicolumn{2}{c}{GT $\geq20$\,m} \\
\cmidrule(lr){2-3}\cmidrule(l){4-5}
Model & $\rho\uparrow$ & $\tau_b\uparrow$ & $\rho\uparrow$ & $\tau_b\uparrow$ \\
\midrule
DAv2-B & 0.463 & 0.338 & 0.084 & 0.060 \\
UniDepth V2-B & 0.425 & 0.307 & 0.076 & 0.055 \\
VGGT & 0.422 & 0.306 & 0.099 & 0.071 \\
CHMv2 & 0.408 & 0.296 & 0.010 & 0.004 \\
VGGT-$\Omega$ & 0.383 & 0.277 & \underline{0.111} & \underline{0.079} \\
MoGe-3-L & 0.379 & 0.272 & 0.080 & 0.056 \\
MetricAnything & 0.366 & 0.262 & 0.085 & 0.061 \\
DAv3-B & 0.186 & 0.130 & 0.029 & 0.019 \\
Metric3D-v2-L & 0.185 & 0.129 & 0.018 & 0.013 \\
RS3DAda & \textbf{0.522} & \textbf{0.385} & 0.096 & 0.068 \\
\textbf{Ours} & \underline{0.488} & \underline{0.365} & \textbf{0.145} & \textbf{0.107} \\
\bottomrule
\end{tabular}%
}
}
\end{minipage}
\end{table*}

\subsection{Main results}

\begin{figure*}[t]
    \centering
    \includegraphics[width=\textwidth]{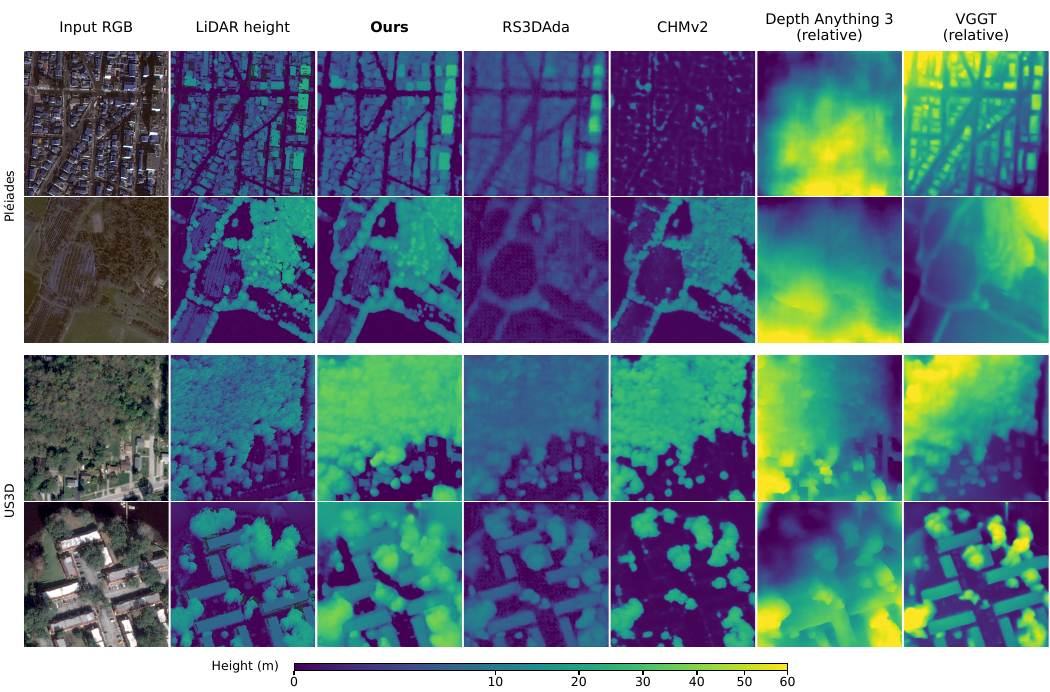}
    
\vspace{-0.8em}
    
    \caption{\textbf{Qualitative height estimation on Pleiades and US3D.}
    Ours denotes our model trained on 150k samples.
    Metric height maps share a 0--60\,m square-root color scale.
    Depth Anything 3 and VGGT use independent relative-depth color scales.
    Predictions are shown without ground-truth alignment.}
    \label{fig:qualitative_comparison}
\end{figure*}

\pparagraph{Monocular height estimation results.}
We benchmark monocular height estimation on our GeomSat test set and on US3D/DFC19~\cite{bosch2019semantic}. MAE scores (in meters) are reported in Table~\ref{tab:height-benchmark}. We compare our model against two families of methods. The first comprises general-purpose geometric models, including Depth Anything V2~\cite{yang2024depthanythingv2}, Depth Anything 3~\cite{lin2025depthanything3}, Metric3D v2~\cite{hu2024metric3dv2}, MoGe~\cite{wang2025moge}, VGGT~\cite{wang2025vggt}, and VGGT-$\Omega$~\cite{wang2026vggtomega}. These methods, which predict relative depth, cannot be directly compared to the ground truth labels. Thus, we follow the methodology used by Ranftl \emph{et al.}~\cite{ranftl2020towards} and Depth Anything~\cite{yang2024depthanything,yang2024depthanythingv2}, fitting one signed scale and offset per image against its ground-truth labels, producing aligned predictions that enable direct comparison. The second family comprises remote-sensing-specific methods that natively output metric height: CHMv2~\cite{brandt2026chmv2}, a global canopy-height specialist, and RS3DAda~\cite{song2024synrs3d}, which combines synthetic SynRS3D pretraining with domain adaptation to US3D~\cite{bosch2019semantic}. We note that on US3D, RS3DAda was trained on crops of the same original satellite products and thus benefits from matched radiometry, illumination, and geography, whereas on GeomSat it faces a genuine cross-sensor transfer. 
On GeomSat, our method outperforms all other approaches in all categories, i.e. overall, non-vegetation, vegetation, and tall pixels $\geq$20\,m. On US3D, the results are mixed: our metric predictions remain competitive for tall vegetation ($\geq$20\,m) but are less accurate on the remaining categories. Since our predictions still look visually coherent, we additionally evaluate our method and CHMv2 under the same per-image alignment used for the general-purpose models. In this setting, our approach becomes more competitive, though the best aligned results are shared across different methods. Fig.~\ref{fig:qualitative_comparison} provides qualitative examples for both datasets on urban and vegetation areas.

\pparagraph{Height ordering and spatial coherence.}
To assess spatial coherency and whether predictions preserve the correct relative structure in the cross sensor case, i.e., that the tallest structure is predicted as tallest, the second tallest as second tallest, and so on, we report rank correlations between predicted and ground-truth height on US3D. Table~\ref{tab:us3d-height-ordering} shows the average per-image Spearman's $\rho$ and Kendall's $\tau_b$ correlations. Our method ranks second overall, behind RS3DAda, which benefits from training on the same exact product acquisitions, and first for tall structures (GT $\geq$20\,m), where $VGGT-\Omega$ scores second. Together with the aligned results in Table~\ref{tab:height-benchmark}, this indicates that even when absolute metric accuracy degrades under cross-sensor transfer, our height predictions remain correctly ordered and spatially coherent.

\pparagraph{Dense correspondence results.}
Dense correspondence is a crucial step in the geometric pipelines of satellite data, such as bundle adjustment and RPC refinement~\cite{mari2021generic, mari2026robust}. Hence, we complement height estimation with this task and benchmark the matchers typically used in such open source pipelines: SIFT~\cite{lowe2004distinctive}, SuperPoint+LightGlue~\cite{sarlin2020superpoint,lindenberger2023lightglue} and XoFTR~\cite{xoftr}. We also add the well-known dense matcher RoMa~\cite{edstedt2024roma} into the comparison. 
Table~\ref{tab:matcher-benchmark} evaluates them on the GeomSat test set. We report three complementary metrics. \emph{Endpoint error} (px, $\downarrow$) is the mean pixel distance between a predicted match and its ground-truth correspondence, and \emph{PCK@3} ($\%$, $\uparrow$) is the fraction of predicted matches whose endpoint error is below 3\,px. Both are computed only on \emph{emitted} matches that can be verified against the ground-truth geometry (RPC cameras), so they measure the \emph{precision} of the matches a method actually produces. \emph{Grid recall@3} ($\%$, $\uparrow$) is the corresponding \emph{coverage} measure: we overlay a regular grid on the source image, restricted to the region verified as visible in the target image, and report the fraction of grid cells that contain at least one verified match with endpoint error below 3\,px. 

\begin{figure*}[t]
    \centering
    \includegraphics[width=\textwidth]{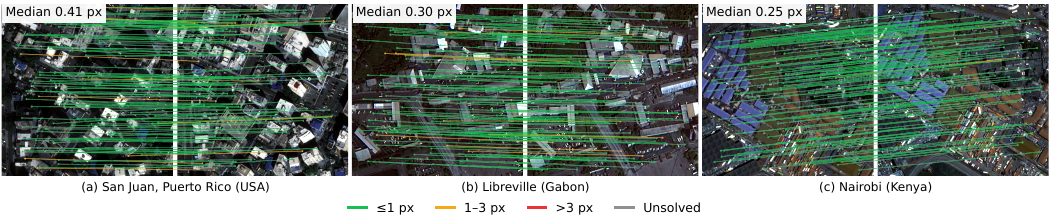}
    
    \vspace{-1.0em}
    
    \caption{\textbf{Correspondences on unseen geographic sites.} For each predicted match, we triangulate a 3D point using bundle-adjusted RPC cameras, then reproject it into both images. Colors indicate the larger of the two pixel distances between the     reprojected point and the predicted endpoints; labels report the median over all predicted matches.     Each panel displays 120 sampled matches without geometric filtering. Low errors indicate geometric consistency with adjusted RPC models, rather than keypoint ground-truth correspondence accuracy.
    }\vspace{-0.5em}
    \label{fig:unseen-world-correspondences}
\end{figure*}

Our method is the most precise with 1.804\,px endpoint error and 87.63\% PCK@3, versus 3.407\,px for SuperPoint+LightGlue and 6.194\,px for XoFTR. RoMa achieves the best coverage (8.49\% grid recall) at lower precision (4.197\,px, 79.06\% PCK). Notice the GeomSat test set includes cross--seasonal examples. Figure~\ref{fig:unseen-world-correspondences} illustrates this geometric coherence on unseen sites: triangulating each predicted match with fixed bundle-adjusted RPC cameras and reprojecting it onto both images yields low pixel error, meaning our correspondences are consistent with standard satellite adjustment pipelines. The examples (San Juan, Libreville, Nairobi) are from the GeomSat test set, so neither pre-training nor fine-tuning saw these geographic areas.

\begin{table}[t]
    \centering
    \caption{Endpoint error and PCK use verified emitted
    matches; grid recall measures coverage of the verified visible source grid. Best results are bold and second-best are underlined.\vspace{-0.7em}}
    \label{tab:matcher-benchmark}
    \renewcommand{\arraystretch}{1.12}
    \scriptsize
    \begin{tabular}{@{}lccc@{}}
        \toprule
        Model
        & \shortstack{Endpoint error\\(px)$\,\downarrow$}
        & \shortstack{PCK@3\\(\%)$\,\uparrow$}
        & \shortstack{Grid recall@3\\(\%)$\,\uparrow$} \\
        \midrule
        XoFTR~\cite{xoftr}
        & 6.194 & 71.23 & 2.54 \\
        SIFT~\cite{lowe2004distinctive}
        & 21.233 & \underline{79.75} & 0.052 \\
        SuperPoint+LightGlue~\cite{sarlin2020superpoint,lindenberger2023lightglue}
        & \underline{3.407} & 66.09 & 0.170 \\
        RoMa-outdoor~\cite{edstedt2024roma} 
        & 4.197 & 79.06 & \textbf{8.49} \\
        \textbf{Ours}
        & \textbf{1.804} & \textbf{87.63} & \underline{3.97} \\
        \bottomrule
    \end{tabular}
    \vspace{-1.5em}
\end{table}

\vspace{-0.25em}
\subsection{Ablation studies}
\label{sec:ablations}
\vspace{-0.5em}
\pparagraph{Pre-training}
Table~\ref{tab:frozen-geometry-transfer} evaluates the transferability of our pre-trained representations to two geometry-related tasks: monocular height estimation (MHE) and building segmentation on WHU~\cite{ji2019whu}, both of which benefit from geometric cues.
For both tasks, we freeze the encoder and train a lightweight
convolutional decoder with a projection to 256
channels and four upsampling stages; architectural details are
provided in the supplementary material.
For MHE, we use 10k training and validation sets images, and approximately 35k test images, all three sets drawn from areas excluded from pre-training. No acquisition metadata is supplied to the height decoder. For building segmentation, we use the aerial subset of the
WHU Building Dataset~\cite{ji2019whu}, which contains
$512\times512$ RGB tiles at 0.3\,m/px resolution with binary building masks. We use its provided splits.

\begin{figure}[t]
    \centering
    \includegraphics[width=\linewidth]{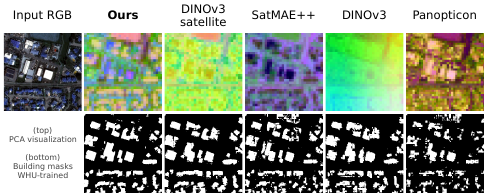}
    \caption{
    \textbf{Top}: feature PCA for an urban scene. 1,000 images from the GeomSat test set are used to fit the PCA for each backbone.
    \textbf{Bottom}: building masks predicted on the scene by
    frozen encoders with the segmentation heads trained
    on WHU. White indicates predicted buildings.\vspace{-1.5em}
    }
    \label{fig:encoder-pca-buildings}
\end{figure}

We compare against CroCo v2~\cite{weinzaepfel2023croco}, DINOv2~\cite{dinov2}, DINOv3~\cite{dinov3} pre-trained on web and satellite imagery, SatMAE++~\cite{satmaepp}, Panopticon~\cite{panopticon}, and AnySat~\cite{anysat}.
AnySat additionally receives near-infrared imagery for MHE and
is thus excluded from RGB-only WHU.
Our encoder achieves the lowest overall and tall-structure
($h\geq20$\,m) height errors, and the highest building IoU, F1,
and mIoU on WHU. It slightly outperforms satellite DINOv3 with significantly fewer parameters and pre-training data volume. Figure~\ref{fig:encoder-pca-buildings} shows the PCA visualization and its predicted building mask for an urban image from the GeomSat test set, illustrating the potential of the learned representations.

\begin{table}[t]
\centering
\caption{\textbf{Frozen-encoder transfer to monocular height
estimation (MHE) on GeomSat and building segmentation on WHU~\cite{ji2019whu}.}
Only task-specific heads are trained.
MHE uses 10k training images without head metadata, measuring
MAE in metres; $\mathrm{MAE}_{20+}$ evaluates pixels
with ground-truth height $\geq20$\,m.
WHU reports building IoU/F1 and background/building mIoU in percent.
AnySat uses RGB+NIR and is not evaluated on RGB-only WHU.\vspace{-.5em}}
\label{tab:frozen-geometry-transfer}
\setlength{\tabcolsep}{4pt}
\renewcommand{\arraystretch}{1.08}
\resizebox{\columnwidth}{!}{%
\begin{tabular}{@{}lcccccc@{}}
\toprule
& & \multicolumn{2}{c}{MHE}
& \multicolumn{3}{c}{WHU Building} \\
\cmidrule(lr){3-4}\cmidrule(lr){5-7}
Encoder & Params (M)
& MAE $\downarrow$ & $\mathrm{MAE}_{20+}$ $\downarrow$
& Build. IoU $\uparrow$ & F1 $\uparrow$ & mIoU $\uparrow$ \\
\midrule
CroCo v2 B/16~\cite{weinzaepfel2023croco}
& 85.6 & 1.48 & 16.51 & 78.17 & 87.75 & 87.60 \\

DINOv2 B/14~\cite{dinov2}
& 86.6 & 1.30 & 13.20 & 79.39 & 88.51 & 88.31 \\

DINOv3 B/16~\cite{dinov3}
& 85.7 & 1.27 & 13.24 & 80.37 & 89.12 & 88.87 \\

DINOv3 sat L/16~\cite{dinov3}
& 303.2 & \underline{1.00} & \underline{9.91}
& \underline{82.81} & \underline{90.60} & \underline{90.26} \\

SatMAE++ L/16~\cite{satmaepp}
& 304.1 & 1.43 & 15.37 & 79.56 & 88.61 & 88.39 \\

Panopticon B/14~\cite{panopticon}
& 99.0 & 1.43 & 16.40 & 70.65 & 82.80 & 83.29 \\

AnySat Base (RGBI)~\cite{anysat}
& 125.9 & 1.61 & 18.59 & --- & --- & --- \\

\textbf{Ours}
& 85.6 & \textbf{0.93} & \textbf{9.34}
& \textbf{82.96} & \textbf{90.69} & \textbf{90.33} \\
\bottomrule
\end{tabular}%
}
\end{table}

\pparagraph{Metadata.}
Acquisition metadata provides information about acquisition geometry and shadows that can be used as height priors. The benefit of metadata is observed in the two top rows of  Table~\ref{tab:height-ablations}: removing metadata increases MAE on tall pixels from 2.92 to 3.35\,m.

\pparagraph{Auxiliary geometric supervision.}
Above-ground displacement fields and surface normals provide complementary geometric signals to direct height, and Table~\ref{tab:height-ablations} also addresses their impact for the MHE task. As shown, above-ground displacement supervision reduces overall MAE from 0.93 to 0.90\,m and tall non-vegetation MAE from 3.35 to 3.19\,m, but increases tall vegetation MAE from 2.90 to 3.27\,m. Normal supervision reduces overall MAE to 0.91\,m, vegetation MAE from 1.69 to 1.67\,m, and tall non-vegetation MAE to 3.14\,m, with a smaller increase in tall vegetation error to 2.97\,m. Neither auxiliary signal improves every population: displacement achieves the lowest overall error among these variants, whereas the normal-supervised model provides a more favorable compromise between vegetation and non-vegetation performance. We thus use the normal-supervised model in our comparisons, despite above-ground displacement fields introduce potentially useful information for non-vegetation areas or other geometric tasks.

\pparagraph{Limitations and future work.}
Despite the geographic diversity of GeomSat, our dataset relies on a single satellite sensor, limiting cross-sensor generalization, as shown by the results in Table~\ref{tab:height-benchmark}. Future work could address this limitation by incorporating imagery from additional satellite sensors or realistic synthetic data to increase sensor diversity. Moreover, our ablations show that above-ground displacement supervision improves height estimation in some settings but degrades performance on tall vegetation. Investigating how to better exploit these geometric cues could yield more consistent improvements.

\begin{table}[t]
    \centering
    \caption{\textbf{Height-estimation ablations.}
    Test MAE (m, $\downarrow$) on 506,015 images after
    fine-tuning on 150k images.
    Tall: GT height $\geq20$\,m; NV/V: non-vegetation/vegetation.
    HM: height with metadata.
    Bold marks the best scores within each block.\vspace{-.5em}}
    \label{tab:height-ablations}
    \setlength{\tabcolsep}{2pt}
    \renewcommand{\arraystretch}{1.02}
    \scriptsize
    \resizebox{\columnwidth}{!}{%
    \begin{tabular}{@{}lrrrrrr@{}}
        \toprule
        & \multicolumn{2}{c}{All}
        & \multicolumn{2}{c}{NV}
        & \multicolumn{2}{c}{V} \\
        \cmidrule(lr){2-3}
        \cmidrule(lr){4-5}
        \cmidrule(lr){6-7}
        Configuration
        & MAE & Tall & MAE & Tall & MAE & Tall \\
        \midrule




        Height, no metadata
        & 0.93 & 3.35
        & 0.49 & 3.92
        & 1.70 & 3.33 \\
        HM
        & 0.93 & \textbf{2.92}
        & 0.50 & 3.35
        & 1.69 & \textbf{2.90} \\
        HM + disp.\ ($\lambda_d{=}0.3$)
        & \textbf{0.90} & 3.27
        & \textbf{0.41} & 3.19
        & 1.76 & 3.27 \\
        HM + normals ($\lambda_n{=}0.1$)
        & 0.91 & 2.98
        & 0.48 & \textbf{3.14}
        & \textbf{1.67} & 2.97 \\
        \bottomrule
    \end{tabular}%
    }\vspace{-0.75em}
\end{table}

\vspace{-0.5em}
\section{Conclusion}
\vspace{-0.25em}
In this work, we introduce a scalable and automated approach to geometric learning from raw VHR satellite imagery. We use Pleiades acquisitions to build a geographically diverse corpus of 1.8 million multi-view samples comprising nearly 9 million images, complemented by dense LiDAR-derived geometric annotations. Through cross-view pre-training and task-specific fine-tuning, we learned representations for monocular height estimation and dense correspondence. Our experiments show strong performance on unseen geographical sites, improving existing methods in metric height estimation and correspondence accuracy. Cross-sensor evaluation shows that although absolute height accuracy degrades under domain shift, the learned representations preserve a meaningful relative height structure, particularly for elevated surfaces. Overall, our work demonstrates the potential of generated multi-view data and geometric supervision for learning transferable satellite representations, providing a foundation for future advances in large-scale 3D scene understanding.

\section{Acknowledgements}
The project was provided with computing HPC and storage resources by GENCI at CINES, thanks to grant AD011016256R1 on the supercomputer Adastra's MI250x partition and at IDRIS, thanks to grants R1001016831R1 and 2023-AD011011801R3.
Centre Borelli is also with Universit\'{e} Paris Cit\'{e}, SSA and INSERM.

\section*{LLM Usage Statement}
We used Codex (OpenAI) and Mistral to conduct research and write this paper. During research, we used them to write code and generate visualizations. During writing, it was used for language editing, improving clarity, and refining sentence formulations..
All scientific claims, experimental results, and sentences were reviewed and validated by the authors.
{
    \small
    \bibliographystyle{ieeenat_fullname}
    \bibliography{main}

@String(PAMI  = {IEEE TPAMI})

@String(IJCV  = {IJCV})

@String(CVPR  = {CVPR})

@String(ICCV  = {ICCV})

@String(ECCV  = {ECCV})

@String(NIPS  = {NeurIPS})

@String(ICLR  = {ICLR})

@String(AAAI = {AAAI})

@String(CVPRW= {CVPRW})

@inproceedings{perez2018film,
  title = {{FiLM}: Visual Reasoning with a General Conditioning Layer},
  author = {Perez, Ethan and Strub, Florian and de Vries, Harm and Dumoulin, Vincent and Courville, Aaron},
  booktitle = {Proceedings of the AAAI Conference on Artificial Intelligence},
  volume = {32},
  number = {1},
  pages = {3942--3951},
  year = {2018},
  doi = {10.1609/aaai.v32i1.11671}
}

@article{weinzaepfel2022croco,
  title={Croco: Self-supervised pre-training for 3d vision tasks by cross-view completion},
  author={Weinzaepfel, Philippe and Leroy, Vincent and Lucas, Thomas and Br{\'e}gier, Romain and Cabon, Yohann and Arora, Vaibhav and Antsfeld, Leonid and Chidlovskii, Boris and Csurka, Gabriela and Revaud, J{\'e}r{\^o}me},
  journal={Advances in Neural Information Processing Systems},
  volume={35},
  pages={3502--3516},
  year={2022}
}

@inproceedings{weinzaepfel2023croco,
  title = {{CroCo v2}: Improved Cross-view Completion Pre-training for Stereo Matching and Optical Flow},
  author = {Weinzaepfel, Philippe and Lucas, Thomas and Leroy, Vincent and Cabon, Yohann and Arora, Vaibhav and Br{\'e}gier, Romain and Csurka, Gabriela and Antsfeld, Leonid and Chidlovskii, Boris and Revaud, J{\'e}r{\^o}me},
  booktitle = ICCV,
  pages = {17969--17980},
  year = {2023},
  doi = {10.1109/ICCV51070.2023.01647}
}

@article{ranftl2020towards,
  title     = {Towards Robust Monocular Depth Estimation: Mixing Datasets for Zero-Shot Cross-Dataset Transfer},
  author    = {Ranftl, Ren{\'e} and Lasinger, Katrin and Hafner, David and Schindler, Konrad and Koltun, Vladlen},
  journal   = {IEEE Transactions on Pattern Analysis and Machine Intelligence},
  volume    = {44},
  number    = {3},
  pages     = {1623--1637},
  year      = {2022},
  doi       = {10.1109/TPAMI.2020.3019967}
}

@inproceedings{ranftl2021vision,
  title     = {Vision Transformers for Dense Prediction},
  author    = {Ranftl, Ren{\'e} and Bochkovskiy, Alexey and Koltun, Vladlen},
  booktitle = {Proceedings of the IEEE/CVF International Conference on Computer Vision (ICCV)},
  pages     = {12179--12188},
  year      = {2021},
  doi       = {10.1109/ICCV48922.2021.01196}
}

@inproceedings{wang2024dust3r,
  title     = {{DUSt3R}: Geometric 3D Vision Made Easy},
  author    = {Wang, Shuzhe and Leroy, Vincent and Cabon, Yohann and Chidlovskii, Boris and Revaud, J{\'e}r{\^o}me},
  booktitle = CVPR,
  pages     = {20697--20709},
  year      = {2024}
}

@article{hu2024metric3dv2,
  title     = {{Metric3D v2}: A Versatile Monocular Geometric Foundation Model for Zero-Shot Metric Depth and Surface Normal Estimation},
  author    = {Hu, Mu and Yin, Wei and Zhang, Chi and Cai, Zhipeng and Long, Xiaoxiao and Chen, Hao and Wang, Kaixuan and Yu, Gang and Shen, Chunhua and Shen, Shaojie},
  journal   = PAMI,
  volume    = {46},
  number    = {12},
  pages     = {10579--10596},
  year      = {2024},
  doi       = {10.1109/TPAMI.2024.3444912}
}

@inproceedings{wang2025vggt,
  title     = {{VGGT}: Visual Geometry Grounded Transformer},
  author    = {Wang, Jianyuan and Chen, Minghao and Karaev, Nikita and Vedaldi, Andrea and Rupprecht, Christian and Novotny, David},
  booktitle = CVPR,
  pages     = {5294--5306},
  year      = {2025}
}

@inproceedings{wang2026vggtomega,
  title = {{VGGT-$\Omega$}},
  author = {Wang, Jianyuan and Chen, Minghao and Zhang, Shangzhan and Karaev, Nikita and Sch{\"o}nberger, Johannes and Labatut, Patrick and Bojanowski, Piotr and Novotny, David and Vedaldi, Andrea and Rupprecht, Christian},
  booktitle = CVPR,
  pages = {21486--21499},
  year = {2026},
  url = {https://openaccess.thecvf.com/content/CVPR2026/html/Wang_VGGT-ohm_CVPR_2026_paper.html}
}

@inproceedings{bosch2019semantic,
  title     = {Semantic Stereo for Incidental Satellite Images},
  author    = {Bosch, Marc and Foster, Kevin and Christie, Gordon and Wang, Sean and Hager, Gregory D. and Brown, Myron},
  booktitle = {IEEE Winter Conference on Applications of Computer Vision},
  pages     = {1524--1532},
  year      = {2019},
  doi       = {10.1109/WACV.2019.00167}
}

@inproceedings{gao2021satmvs,
  title     = {Rational Polynomial Camera Model Warping for Deep Learning Based Satellite Multi-View Stereo Matching},
  author    = {Gao, Jian and Liu, Jin and Ji, Shunping},
  booktitle = ICCV,
  pages     = {6148--6157},
  year      = {2021},
  doi       = {10.1109/ICCV48922.2021.00609}
}

@inproceedings{masquil2025seo,
  title     = {{S-EO}: A Large-Scale Dataset for Geometry-Aware Shadow Detection in Remote Sensing Applications},
  author    = {Masquil, El{\'\i}as and Mar{\'\i}, Roger and Ehret, Thibaud and Meinhardt-Llopis, Enric and Mus{\'e}, Pablo and Facciolo, Gabriele},
  booktitle = CVPRW,
  pages     = {2408--2418},
  year      = {2025},
  url       = {https://openaccess.thecvf.com/content/CVPR2025W/EarthVision/html/Masquil_S-EO_A_Large-Scale_Dataset_for_Geometry-Aware_Shadow_Detection_in_Remote_CVPRW_2025_paper.html}
}

@article{zhu2025globalbuildingatlas,
  title     = {{GlobalBuildingAtlas}: An Open Global and Complete Dataset of Building Polygons, Heights and {LoD1} 3D Models},
  author    = {Zhu, Xiao Xiang and Chen, Sining and Zhang, Fahong and Shi, Yilei and Wang, Yuanyuan},
  journal   = {Earth System Science Data},
  volume    = {17},
  number    = {12},
  pages     = {6647--6668},
  year      = {2025},
  doi       = {10.5194/essd-17-6647-2025}
}

@inproceedings{yang2024depthanythingv2,
  title     = {Depth Anything V2},
  author    = {Yang, Lihe and Kang, Bingyi and Huang, Zilong and Zhao, Zhen and Xu, Xiaogang and Feng, Jiashi and Zhao, Hengshuang},
  booktitle = NIPS,
  volume    = {37},
  year      = {2024},
  doi       = {10.52202/079017-0688}
}

@inproceedings{derksen2021snerf,
  title = {Shadow Neural Radiance Fields for Multi-View Satellite Photogrammetry},
  author = {Derksen, Dawa and Izzo, Dario},
  booktitle = CVPRW,
  pages = {1152--1161},
  year = {2021},
  url = {https://openaccess.thecvf.com/content/CVPR2021W/EarthVision/html/Derksen_Shadow_Neural_Radiance_Fields_for_Multi-View_Satellite_Photogrammetry_CVPRW_2021_paper.html}
}

@inproceedings{mari2022satnerf,
  title = {{Sat-NeRF}: Learning Multi-View Satellite Photogrammetry With Transient Objects and Shadow Modeling Using {RPC} Cameras},
  author = {Mar{\'\i}, Roger and Facciolo, Gabriele and Ehret, Thibaud},
  booktitle = CVPRW,
  pages = {1311--1321},
  year = {2022},
  doi = {10.1109/CVPRW56347.2022.00137}
}

@inproceedings{mari2023eonerf,
  title = {Multi-Date Earth Observation {NeRF}: The Detail Is in the Shadows},
  author = {Mar{\'\i}, Roger and Facciolo, Gabriele and Ehret, Thibaud},
  booktitle = CVPRW,
  pages = {2035--2045},
  year = {2023},
  url = {https://openaccess.thecvf.com/content/CVPR2023W/EarthVision/html/Mari_Multi-Date_Earth_Observation_NeRF_The_Detail_Is_in_the_Shadows_CVPRW_2023_paper.html}
}

@inproceedings{li2018megadepth,
  title = {{MegaDepth}: Learning Single-View Depth Prediction From Internet Photos},
  author = {Li, Zhengqi and Snavely, Noah},
  booktitle = CVPR,
  pages = {2041--2050},
  year = {2018},
  url = {https://openaccess.thecvf.com/content_cvpr_2018/html/Li_MegaDepth_Learning_Single-View_CVPR_2018_paper.html}
}

@inproceedings{yang2024depthanything,
  title = {Depth Anything: Unleashing the Power of Large-Scale Unlabeled Data},
  author = {Yang, Lihe and Kang, Bingyi and Huang, Zilong and Xu, Xiaogang and Feng, Jiashi and Zhao, Hengshuang},
  booktitle = CVPR,
  pages = {10371--10381},
  year = {2024},
  doi = {10.1109/CVPR52733.2024.00987}
}

@inproceedings{lin2025depthanything3,
  title = {Depth Anything 3: Recovering the Visual Space from Any Views},
  author = {Lin, Haotong and Chen, Sili and Liew, Jun Hao and Chen, Donny Y. and Li, Zhenyu and Zhao, Yang and Peng, Sida and Guo, Hengkai and Zhou, Xiaowei and Shi, Guang and Feng, Jiashi and Kang, Bingyi},
  booktitle = ICLR,
  year = {2026},
  url = {https://proceedings.iclr.cc/paper_files/paper/2026/hash/e4cd50120b6d7e8daff1749d6bbaa889-Abstract-Conference.html}
}

@article{mari2021generic,
  title={A generic bundle adjustment methodology for indirect RPC model refinement of satellite imagery},
  author={Mar{\'\i}, Roger and de Franchis, Carlo and Meinhardt-Llopis, Enric and Anger, J{\'e}r{\'e}my and Facciolo, Gabriele},
  journal={Image Processing On Line},
  volume={11},
  pages={344--373},
  year={2021},
  doi={10.5201/ipol.2021.352}
}

@inproceedings{mari2019bundle,
  title = {To Bundle Adjust or Not: A Comparison of Relative Geolocation Correction Strategies for Satellite Multi-View Stereo},
  author = {Mar{\'\i}, Roger and de Franchis, Carlo and Meinhardt-Llopis, Enric and Facciolo, Gabriele},
  booktitle = {Proceedings of the IEEE/CVF International Conference on Computer Vision (ICCV) Workshops},
  year = {2019},
  url = {https://openaccess.thecvf.com/content_ICCVW_2019/html/3DRW/Mari_To_Bundle_Adjust_or_Not_A_Comparison_of_Relative_Geolocation_ICCVW_2019_paper.html}
}

@inproceedings{mari2026robust,
  title     = {Robust RPC Bundle Adjustment for Multi-Date Satellite
               Imagery with Season-Invariant Correspondences},
  author    = {Marí, Roger and Masquil, Elías and
               Bou, Xavier and Ehret, Thibaud and
               Facciolo, Gabriele},
  booktitle = {TerraBytes II: ECCV 2026 Workshop},
  year      = {2026}
}

@inproceedings{wang2025moge,
  title     = {{MoGe}: Unlocking Accurate Monocular Geometry Estimation for Open-Domain Images with Optimal Training Supervision},
  author    = {Wang, Ruicheng and Xu, Sicheng and Dai, Cassie and Xiang, Jianfeng and Deng, Yu and Tong, Xin and Yang, Jiaolong},
  booktitle = {Proceedings of the IEEE/CVF Conference on Computer Vision and Pattern Recognition (CVPR)},
  pages     = {5261--5271},
  year      = {2025}
}

@inproceedings{leroy2024mast3r,
  title = {Grounding Image Matching in 3D with {MASt3R}},
  author = {Leroy, Vincent and Cabon, Yohann and Revaud, J{\'e}r{\^o}me},
  booktitle = ECCV,
  year = {2024},
  url = {https://www.ecva.net/papers/eccv_2024/papers_ECCV/html/9080_ECCV_2024_paper.php}
}

@inproceedings{huang2026skysplat,
  title     = {{SkySplat}: Generalizable 3D Gaussian Splatting from Multi-Temporal Sparse Satellite Images},
  author    = {Huang, Xuejun and Liu, Xinyi and Wan, Yi and Zheng, Zhi and Zhang, Bin and Xiong, Mingtao and Pei, Yingying and Zhang, Yongjun},
  booktitle = {Proceedings of the AAAI Conference on Artificial Intelligence},
  volume    = {40},
  number    = {7},
  pages     = {5158--5166},
  year      = {2026},
  doi       = {10.1609/aaai.v40i7.37430}
}

@article{luo2026eovggt,
  title         = {{EO-VGGT}: Orbital Ray-Conditioned 3D Foundation Models for Satellite Multi-View Reconstruction},
  author        = {Luo, Qiyan and Pi, Yingdong and Wen, Lekang and Yang, Jie and Wang, Xiaoyu and Zhang, Haiming and Wang, Mi},
  journal       = {arXiv preprint arXiv:2607.00417},
  year          = {2026},
  eprint        = {2607.00417},
  archivePrefix = {arXiv}
}

@article{chen2023htcdc,
  title   = {{HTC-DC Net}: Monocular Height Estimation From Single Remote Sensing Images},
  author  = {Chen, Sining and Shi, Yilei and Xiong, Zhitong and Zhu, Xiao Xiang},
  journal = {IEEE Transactions on Geoscience and Remote Sensing},
  volume  = {61},
  pages   = {1--18},
  year    = {2023},
  doi     = {10.1109/TGRS.2023.3321255}
}

@article{xiong2023the,
  title   = {{THE} Benchmark: Transferable Representation Learning for Monocular Height Estimation},
  author  = {Xiong, Zhitong and Huang, Wei and Hu, Jingtao and Zhu, Xiao Xiang},
  journal = {IEEE Transactions on Geoscience and Remote Sensing},
  volume  = {61},
  pages   = {1--14},
  year    = {2023},
  doi     = {10.1109/TGRS.2023.3311764}
}

@inproceedings{song2024synrs3d,
  title = {{SynRS3D}: A Synthetic Dataset for Global 3D Semantic Understanding from Monocular Remote Sensing Imagery},
  author = {Song, Jian and Chen, Hongruixuan and Xuan, Weihao and Xia, Junshi and Yokoya, Naoto},
  booktitle = {Advances in Neural Information Processing Systems, Datasets and Benchmarks Track},
  volume = {37},
  year = {2024},
  doi = {10.52202/079017-3727}
}

@article{jiang2026rdah,
  title   = {{RDAH-Net}: Bridging Relative Depth and Absolute Height for Monocular Height Estimation in Remote Sensing},
  author  = {Jiang, Liting and Wang, Feng and Jiao, Niangang and Zhu, Jingxing and Xiang, Yuming and You, Hongjian},
  journal = {Remote Sensing},
  volume  = {18},
  number  = {7},
  pages   = {1024},
  year    = {2026},
  doi     = {10.3390/rs18071024}
}

@article{brandt2026chmv2,
  title   = {{CHMv2}: Improvements in Global Canopy Height Mapping using {DINOv3}},
  author  = {Brandt, John and Yi, Seungeun and Tolan, Jamie and Li, Xinyuan and Potapov, Peter and Ertel, Jessica and Spore, Justine and Vo, Huy V. and Ramamonjisoa, Micha{\"e}l and Labatut, Patrick and Bojanowski, Piotr and Couprie, Camille},
  journal = {Scientific Data},
  year    = {2026},
  doi     = {10.1038/s41597-026-07882-0}
}

@inproceedings{christie2020geocentric,
  title     = {Learning Geocentric Object Pose in Oblique Monocular Images},
  author    = {Christie, Gordon and Abujder, Rodrigo Rene Rai Munoz and Foster, Kevin and Hagstrom, Shea and Hager, Gregory D. and Brown, Myron Z.},
  booktitle = {Proceedings of the IEEE/CVF Conference on Computer Vision and Pattern Recognition (CVPR)},
  pages     = {14512--14520},
  year      = {2020}
}

@inproceedings{li2021monocularbuilding,
  title     = {3D Building Reconstruction From Monocular Remote Sensing Images},
  author    = {Li, Weijia and Meng, Lingxuan and Wang, Jinwang and He, Conghui and Xia, Gui-Song and Lin, Dahua},
  booktitle = {Proceedings of the IEEE/CVF International Conference on Computer Vision (ICCV)},
  pages     = {12548--12557},
  year      = {2021}
}

@inproceedings{li2024multilevelbuilding,
  title     = {3D Building Reconstruction from Monocular Remote Sensing Images with Multi-level Supervisions},
  author    = {Li, Weijia and Yang, Haote and Hu, Zhenghao and Zheng, Juepeng and Xia, Gui-Song and He, Conghui},
  booktitle = {Proceedings of the IEEE/CVF Conference on Computer Vision and Pattern Recognition (CVPR)},
  pages     = {27728--27737},
  year      = {2024}
}

@inproceedings{cong2022satmae,
  title     = {{SatMAE}: Pre-training Transformers for Temporal and Multi-Spectral Satellite Imagery},
  author    = {Cong, Yezhen and Khanna, Samar and Meng, Chenlin and Liu, Patrick and Rozi, Erik and He, Yutong and Burke, Marshall and Lobell, David and Ermon, Stefano},
  booktitle = {Advances in Neural Information Processing Systems},
  volume    = {35},
  year      = {2022}
}

@inproceedings{reed2023scalemae,
  title     = {{Scale-MAE}: A Scale-Aware Masked Autoencoder for Multiscale Geospatial Representation Learning},
  author    = {Reed, Colorado J. and Gupta, Ritwik and Li, Shufan and Brockman, Sarah and Funk, Christopher and Clipp, Brian and Keutzer, Kurt and Candido, Salvatore and Uyttendaele, Matt and Darrell, Trevor},
  booktitle = {Proceedings of the IEEE/CVF International Conference on Computer Vision (ICCV)},
  pages     = {4088--4099},
  year      = {2023}
}

@inproceedings{guo2024skysense,
  title     = {{SkySense}: A Multi-Modal Remote Sensing Foundation Model Towards Universal Interpretation for Earth Observation Imagery},
  author    = {Guo, Xin and Lao, Jiangwei and Dang, Bo and Zhang, Yingying and Yu, Lei and Ru, Lixiang and Zhong, Liheng and Huang, Ziyuan and Wu, Kang and Hu, Dingxiang and He, Huimei and Wang, Jian and Chen, Jingdong and Yang, Ming and Zhang, Yongjun and Li, Yansheng},
  booktitle = {Proceedings of the IEEE/CVF Conference on Computer Vision and Pattern Recognition (CVPR)},
  pages     = {27672--27683},
  year      = {2024}
}

@article{mutreja2026hires,
  title   = {Disentangling Geometry and Texture: An Investigation into Large-Scale {RGB-DSM} Pre-training for 3D Urban Understanding},
  author  = {Mutreja, Guneet and Bittner, Ksenia},
  journal = {ISPRS Open Journal of Photogrammetry and Remote Sensing},
  volume  = {21},
  pages   = {100139},
  year    = {2026},
  doi     = {10.1016/j.ophoto.2026.100139}
}

@inproceedings{eigen2015predicting,
  author    = {Eigen, David and Fergus, Rob},
  title     = {Predicting Depth, Surface Normals and Semantic Labels
               with a Common Multi-Scale Convolutional Architecture},
  booktitle = {Proceedings of the IEEE International Conference
               on Computer Vision (ICCV)},
  year      = {2015},
  pages     = {2650--2658}
}

@inproceedings{xoftr,
  author    = {Tuzcuo\u{g}lu, \"Onder and K\"oksal, Aybora
               and Sofu, Bu\u{g}ra and Kalkan, Sinan
               and Alatan, A. Aydin},
  title     = {{XoFTR}: Cross-modal Feature Matching Transformer},
  booktitle = {Proceedings of the IEEE/CVF Conference on Computer
               Vision and Pattern Recognition (CVPR) Workshops},
  year      = {2024},
  pages     = {4275--4286}
}

@article{lowe2004distinctive,
  title     = {Distinctive Image Features from Scale-Invariant Keypoints},
  author    = {Lowe, David G.},
  journal   = IJCV,
  volume    = {60},
  number    = {2},
  pages     = {91--110},
  year      = {2004},
  doi       = {10.1023/B:VISI.0000029664.99615.94}
}

@inproceedings{sarlin2020superpoint,
  title = {{SuperPoint}: Self-Supervised Interest Point Detection and Description},
  author = {DeTone, Daniel and Malisiewicz, Tomasz and Rabinovich, Andrew},
  booktitle = {Proceedings of the IEEE Conference on Computer Vision and Pattern Recognition (CVPR) Workshops},
  pages = {224--236},
  year = {2018},
  url = {https://openaccess.thecvf.com/content_cvpr_2018_workshops/w9/html/DeTone_SuperPoint_Self-Supervised_Interest_CVPR_2018_paper.html}
}

@inproceedings{lindenberger2023lightglue,
  title = {{LightGlue}: Local Feature Matching at Light Speed},
  author = {Lindenberger, Philipp and Sarlin, Paul-Edouard and Pollefeys, Marc},
  booktitle = ICCV,
  pages = {17627--17638},
  year = {2023},
  doi = {10.1109/ICCV51070.2023.01616}
}

@inproceedings{edstedt2024roma,
  title = {{RoMa}: Robust Dense Feature Matching},
  author = {Edstedt, Johan and Sun, Qiyu and B{\"o}kman, Georg and Wadenb{\"a}ck, M{\aa}rten and Felsberg, Michael},
  booktitle = CVPR,
  pages = {19790--19800},
  year = {2024},
  url = {https://openaccess.thecvf.com/content/CVPR2024/html/Edstedt_RoMa_Robust_Dense_Feature_Matching_CVPR_2024_paper.html}
}

@misc{pleiades_product_guide,
  title = {Pl{\'e}iades Imagery User Guide},
  author = {{Airbus Defence and Space Intelligence}},
  year = {2012},
  howpublished = {Technical reference guide, document USRPHR-DT-125-SPOT-2.0},
  url = {https://www.rymdstyrelsen.se/contentassets/de0a9d767f6644389feab11535c91530/airbus-pleiades-imagery-user-guide-15042021.pdf},
  note = {Document control date: October 18, 2012; file distributed by a third-party mirror}
}

@misc{ign_lidarhd,
  title        = {Lidar HD -- Couverture lidar altim\'etrique haute densit\'e de la France m\'etropolitaine},
  author       = {{Institut national de l'information g\'eographique et foresti\`ere (IGN)}},
  year         = {2024},
  howpublished = {\url{https://geoservices.ign.fr/documentation/donnees/lidarhd}},
  note         = {Accessed: 2026}
}

@misc{usgs_3d,
  title = {{3D Elevation Program (3DEP)}},
  author = {{U.S. Geological Survey}},
  year = {2026},
  howpublished = {U.S. Geological Survey (USGS)},
  url = {https://www.usgs.gov/3d-elevation-program},
  note = {Accessed: 2026}
}

@InProceedings{Zhang_2026_CVPR,
    author    = {Zhang, Chuhan and Le Moing, Guillaume and Koppula, Skanda and Rocco, Ignacio and Momeni, Liliane and Xie, Junyu and Sun, Shuyang and Sukthankar, Rahul and Barral, Jo\"elle K. and Hadsell, Raia and Ghahramani, Zoubin and Zisserman, Andrew and Zhang, Junlin and Sajjadi, Mehdi S. M.},
    title     = {Efficiently Reconstructing Dynamic Scenes One D4RT at a Time},
    booktitle = {Proceedings of the IEEE/CVF Conference on Computer Vision and Pattern Recognition (CVPR)},
    month     = {6},
    year      = {2026},
    pages     = {7382-7392}
}

@inproceedings{loiseau2025alligat0r,
  title = {{Alligat0R}: Pre-Training through Covisibility Segmentation for Relative Camera Pose Regression},
  author = {Loiseau, Thibaut and Bourmaud, Guillaume and Lepetit, Vincent},
  booktitle = NIPS,
  volume = {38},
  year = {2025},
  doi = {10.52202/085713-0463}
}

@inproceedings{diaz2026seasonstereo,
  title = {{SeasonStereo}: Robust Dense Stereo Matching for Multi-Date Satellite Imagery via Generative AI},
  author = {D{\'\i}az-Laureano, {\'A}lvaro and Mar{\'\i}, Roger and Masquil, El{\'\i}as and Arias, Pablo and Facciolo, Gabriele},
  booktitle = {TerraBytes II Workshop at ECCV},
  year = {2026},
  url = {https://github.com/Alvaritox11/SeasonStereo}
}

@article{masquil2026diachronic,
  title = {Diachronic Stereo Matching for Multi-Date Satellite Imagery},
  author = {Masquil, El{\'\i}as and Aira, Luca Savant and Mar{\'\i}, Roger and Ehret, Thibaud and Mus{\'e}, Pablo and Facciolo, Gabriele},
  journal = {ISPRS Annals of the Photogrammetry, Remote Sensing and Spatial Information Sciences},
  volume = {XI-2-2026},
  pages = {483--492},
  year = {2026},
  doi = {10.5194/isprs-annals-XI-2-2026-483-2026}
}

@inproceedings{aira2025gaussian,
  title={Gaussian splatting for efficient satellite image photogrammetry},
  author={Aira, Luca Savant and Facciolo, Gabriele and Ehret, Thibaud},
  booktitle={2025 IEEE/CVF Conference on Computer Vision and Pattern Recognition (CVPR)},
  pages={5959--5969},
  year={2025},
  organization={IEEE}
}

@article{ji2019whu,
  author  = {Ji, Shunping and Wei, Shiqing and Lu, Meng},
  title   = {Fully Convolutional Networks for Multisource Building Extraction From an Open Aerial and Satellite Imagery Data Set},
  journal = {IEEE Transactions on Geoscience and Remote Sensing},
  year    = {2019},
  volume  = {57},
  number  = {1},
  pages   = {574--586},
  doi     = {10.1109/TGRS.2018.2858817}
}

@article{dinov2,
  title = {{DINOv2}: Learning Robust Visual Features without Supervision},
  author = {Oquab, Maxime and Darcet, Timoth{\'e}e and Moutakanni, Th{\'e}o and Vo, Huy V. and Szafraniec, Marc and Khalidov, Vasil and Fernandez, Pierre and Haziza, Daniel and Massa, Francisco and El-Nouby, Alaaeldin and Assran, Mido and Ballas, Nicolas and Galuba, Wojciech and Howes, Russell and Huang, Po-Yao and Li, Shang-Wen and Misra, Ishan and Rabbat, Michael and Sharma, Vasu and Synnaeve, Gabriel and Xu, Hu and J{\'e}gou, Herv{\'e} and Mairal, Julien and Labatut, Patrick and Joulin, Armand and Bojanowski, Piotr},
  journal = {Transactions on Machine Learning Research},
  year = {2024},
  url = {https://mlanthology.org/tmlr/2024/oquab2024tmlr-dinov2/}
}

@article{dinov3,
  title = {{DINOv3}},
  author = {Sim{\'e}oni, Oriane and Vo, Huy V. and Seitzer, Maximilian and Baldassarre, Federico and Oquab, Maxime and Jose, Cijo and Khalidov, Vasil and Szafraniec, Marc and Yi, Seung Eun and Ramamonjisoa, Michael and Massa, Francisco and Haziza, Daniel and Wehrstedt, Luca and Wang, Jianyuan and Darcet, Timoth{\'e}e and Moutakanni, Th{\'e}o and Sentana, Leonel and Roberts, Claire and Vedaldi, Andrea and Tolan, Jamie and Brandt, John and Couprie, Camille and Mairal, Julien and J{\'e}gou, Herv{\'e} and Labatut, Patrick and Bojanowski, Piotr},
  journal = {Transactions on Machine Learning Research},
  year = {2026},
  url = {https://mlanthology.org/tmlr/2026/simeoni2026tmlr-dinov3/}
}

@inproceedings{satmaepp,
  title     = {Rethinking Transformers Pre-training for Multi-Spectral Satellite Imagery},
  author    = {Noman, Mubashir and Naseer, Muzammal and Cholakkal, Hisham and Anwer, Rao Muhammad and Khan, Salman and Khan, Fahad Shahbaz},
  booktitle = CVPR,
  pages     = {27811--27819},
  year      = {2024}
}

@inproceedings{panopticon,
  title     = {{Panopticon}: Advancing Any-Sensor Foundation Models for Earth Observation},
  author    = {Waldmann, Leonard and Shah, Ando and Wang, Yi and Lehmann, Nils and Stewart, Adam and Xiong, Zhitong and Zhu, Xiao Xiang and Bauer, Stefan and Chuang, John},
  booktitle = CVPRW,
  pages     = {2229--2239},
  year      = {2025}
}

@inproceedings{anysat,
  title     = {{AnySat}: One Earth Observation Model for Many Resolutions, Scales, and Modalities},
  author    = {Astruc, Guillaume and Gonthier, Nicolas and Mallet, Cl{\'e}ment and Landrieu, Loic},
  booktitle = CVPR,
  pages     = {19530--19540},
  year      = {2025}
}

@article{gibril2024transformer,
  author  = {Gibril, Mohamed Barakat A. and Al-Ruzouq, Rami and
             Shanableh, Abdallah and Jena, Ratiranjan and Bolcek, Jan and
             Shafri, Helmi Zulhaidi Mohd and Ghorbanzadeh, Omid},
  title   = {Transformer-based semantic segmentation for large-scale
             building footprint extraction from very-high resolution
             satellite images},
  journal = {Advances in Space Research},
  year    = {2024},
  volume  = {73},
  number  = {10},
  pages   = {4937--4954},
  doi     = {10.1016/j.asr.2024.03.002}
}

@article{tucker2023carbon,
  author  = {Tucker, Compton and Brandt, Martin and Hiernaux, Pierre and
             Kariryaa, Ankit and Rasmussen, Kjeld and Small, Jennifer and
             Igel, Christian and Reiner, Florian and Melocik, Katherine and
             Meyer, Jesse and Sinno, Scott and Romero, Eric and
             Glennie, Erin and Fitts, Yaxuan and Morin, Annelise and
             Pinzon, Jorge and McClain, David and Morin, Paul and
             Porter, Claire and Loeffler, Shane and Kergoat, Laurent and
             Issoufou, Bil-Assanou and Savadogo, Patrice and
             Wigneron, Jean-Pierre and Poulter, Benjamin and Ciais, Philippe and
             Kaufmann, Robert and Myneni, Ranga and Saatchi, Sassan and
             Fensholt, Rasmus},
  title   = {Sub-continental-scale carbon stocks of individual trees
             in African drylands},
  journal = {Nature},
  year    = {2023},
  volume  = {615},
  pages   = {80--86},
  doi     = {10.1038/s41586-022-05653-6}
}

@article{workman2023prediction,
  author  = {Workman, Robin and Wong, Patrick and Wright, Alex and Wang, Zhao},
  title   = {Prediction of Unpaved Road Conditions Using High-Resolution
             Optical Satellite Imagery and Machine Learning},
  journal = {Remote Sensing},
  year    = {2023},
  volume  = {15},
  number  = {16},
  pages   = {3985},
  doi     = {10.3390/rs15163985}
}

@article{hafner2025disasteradaptivenet,
  author  = {Hafner, Sebastian and Gerard, Sebastien and
             Sullivan, Josephine and Ban, Yifang},
  title   = {{DisasterAdaptiveNet}: A robust network for multi-hazard
             building damage detection from very-high-resolution
             satellite imagery},
  journal = {International Journal of Applied Earth Observation and Geoinformation},
  year    = {2025},
  volume  = {143},
  pages   = {104756},
  doi     = {10.1016/j.jag.2025.104756}
}
}
\clearpage
\setcounter{page}{1}
\maketitlesupplementary
\appendix

This supplementary material complements the main text with additional details and experiments.

\section{General}
\begin{figure}[t]
    \centering
    \includegraphics[
        width=\columnwidth,
        trim=0 0 0 0,
        clip
    ]{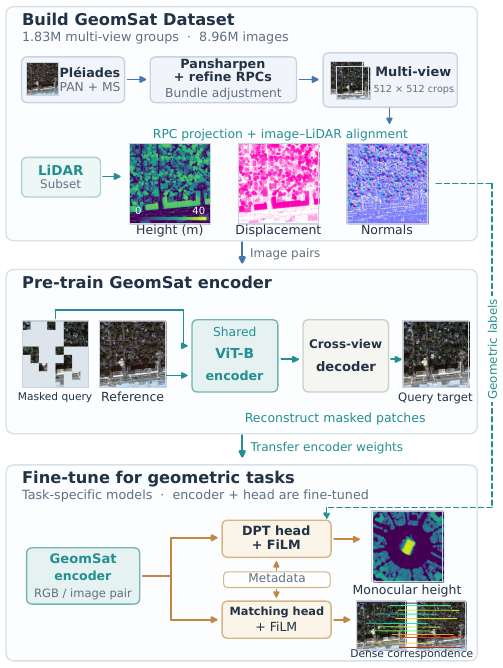}
    \caption{
        \textbf{General diagram of the proposed GeomSat methodology.}
        We process raw data to generate a geographically distributed multi-view dataset with geometric supervision labels where possible. Then, we pre-train an encoder via cross-view completion using these multi-view samples. Lastly, we leverage the geometric labels to fine-tune the learned representations for geometric tasks.
        \vspace{-1em}
    }
    \label{fig:method_diagram}
\end{figure}

\begin{figure*}[t]
    \centering
    \includegraphics[
        width=\textwidth,
        trim=20 38 20 30,
        clip
    ]{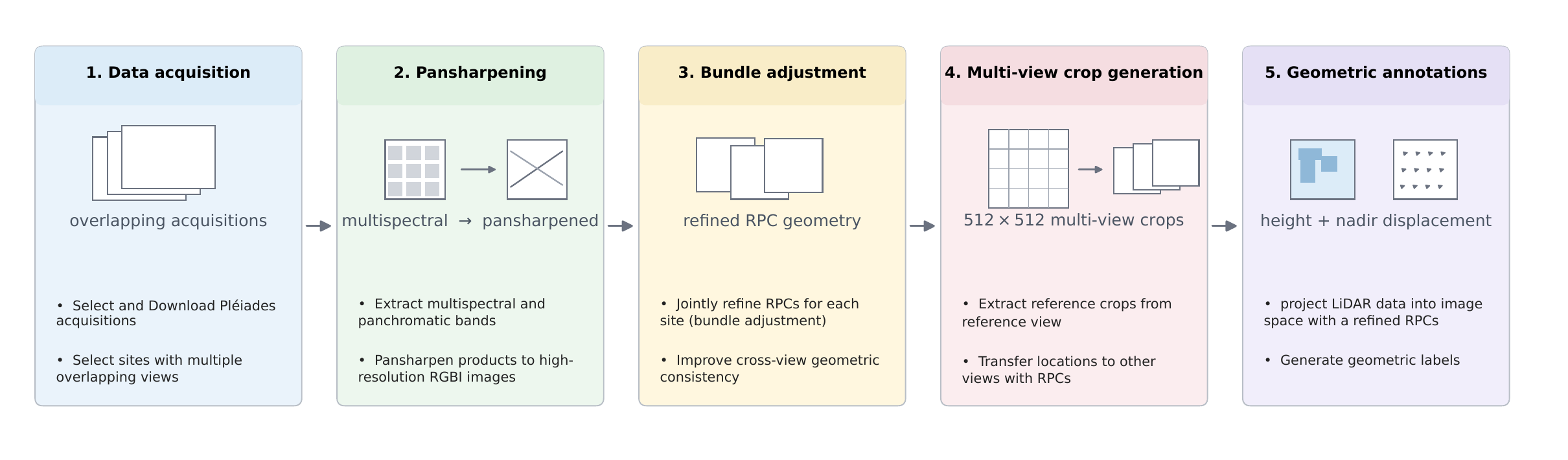}
    \caption{
        \textbf{GeomSat dataset processing pipeline.} From raw products, we process the data to generate ready-to-train geometric samples. This includes pansharpening, RPC refinement via bundle adjustment, multi-view crop generation, and geometric label generation. 
    }
    \label{fig:dataset_pipeline}
\end{figure*}

In this section we provide two graphical representations of the proposed work. First, Figure~\ref{fig:method_diagram} illustrates the general diagram of our work where, from raw satellite products, we derive a geographically distributed dataset with geometric supervision labels, pre-train an encoder on such data, and later fine-tune the learned representations for geometric-dependent tasks.
Secondly, Figure~\ref{fig:dataset_pipeline} shows the data processing steps to generate the GeomSat dataset, including data acquisition, pansharpening, RPC refinement via bundle adjustment, multi-view crop generation with a reference product, and the label generation process.

\section{The GeomSat dataset}

\begin{figure*}[t]
    \centering
    \includegraphics[width=\textwidth]{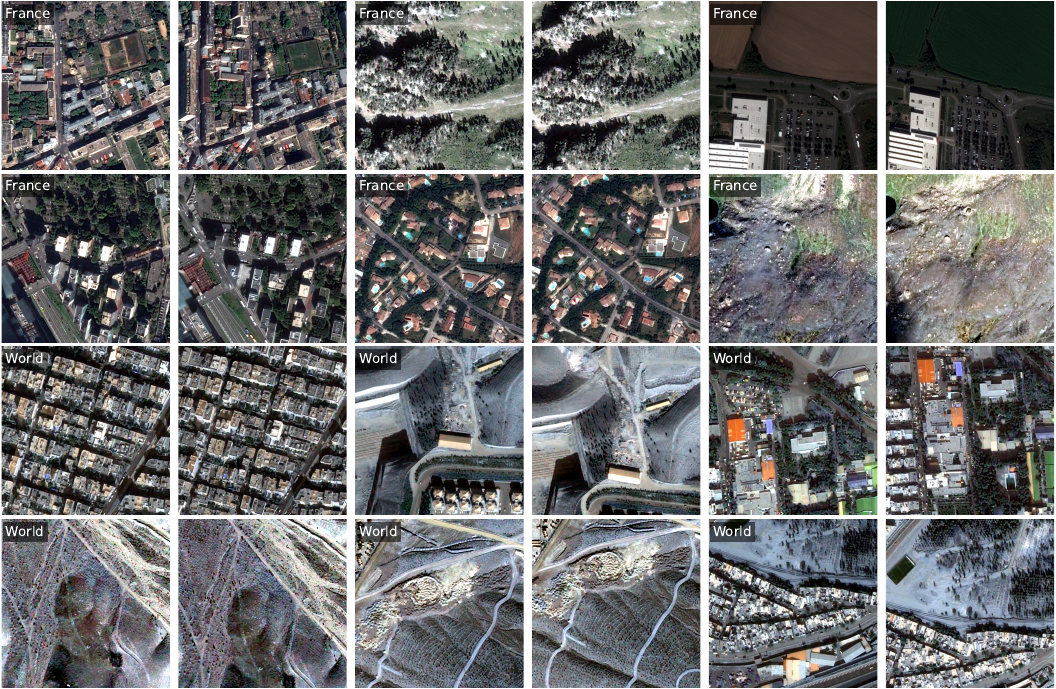}
    \caption{
        \textbf{Multi-view pair examples of the GeomSat dataset.} Some observations correspond to areas in France, where Pleiades acquisitions are abundant, while other correspond to other areas of the World.
    }
    \label{fig:dataset_example_pairs}
\end{figure*}

\begin{figure*}[t]
    \centering
    \includegraphics[
        width=\textwidth,
        trim=0 0 0 0,
        clip
    ]{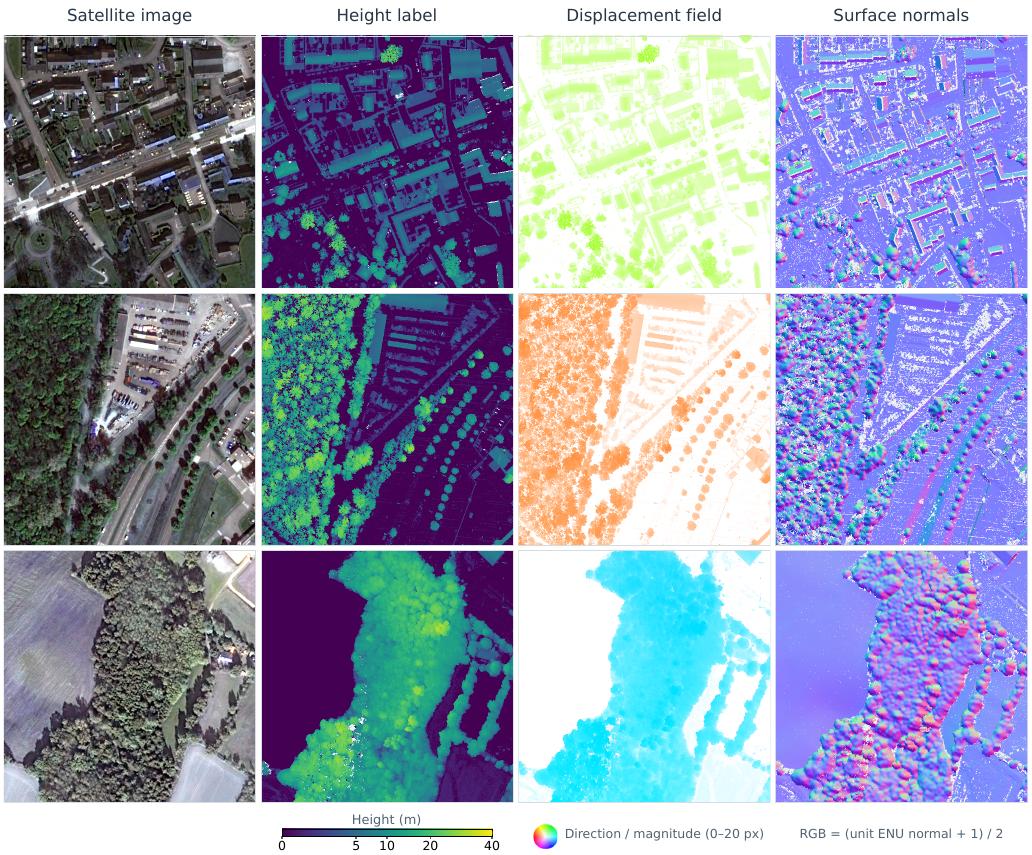}
    \caption{
        \textbf{Examples of GeomSat geometric labels.}
        We show input image, height label, above-height displacement field and normal field annotations.
    }
    \label{fig:geometric_labels}
\end{figure*}

\pparagraph{Dataset examples}
We provide in Figure~\ref{fig:dataset_example_pairs} an example of the generated multi-view observations used during the pre-training. They include examples in France, where Pleiades acquisitions are abundant, as well as other areas of the World. Moreover, in Figure~\ref{fig:geometric_labels} we show examples of the generated GeomSat geometric labels, including height map, above-height displacement field and normals for three input RGB examples.

\begin{figure}[t]
    \centering
    \includegraphics[
        width=\columnwidth,
        trim=0 0 0 30,
        clip
    ]{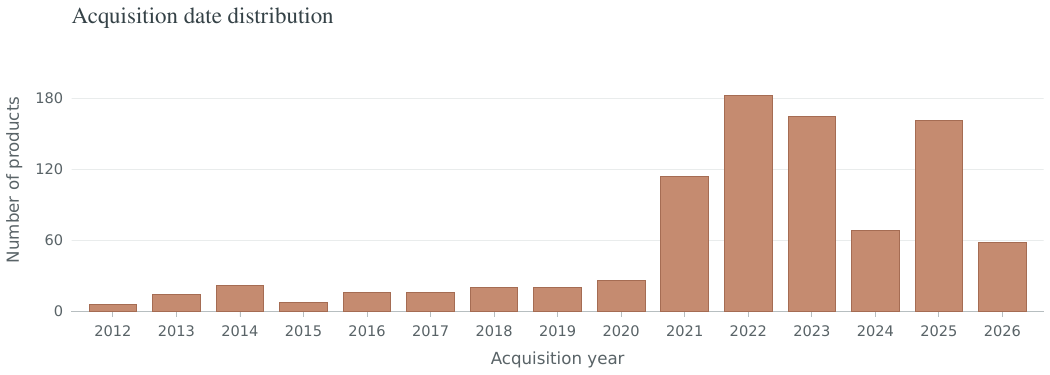}
    \caption{
        \textbf{Acquisition date distribution on the GeomSat dataset.}
        Most acquisitions correspond to the 2021-2026 period, with some observations in the 2012-2020 range as well.
    }
    \label{fig:dataset_date_distribution}
\end{figure}
\pparagraph{Comparison to previous works.}
Generic geometric foundation models have shown the value of pretraining at scale: CroCo-v2 uses \(5.3\)M real cross-view pairs, DUSt3R \(8.5\)M heterogeneous pairs, and Metric3D-v2 more than \(16\)M geometrically annotated images~\cite{weinzaepfel2023croco,wang2024dust3r,hu2024metric3dv2}. VGGT and VGGT-\(\Omega\) further move toward large-scale scene-level multi-view learning~\cite{wang2025vggt,wang2026vggtomega}. Our goal is not to exceed these generic corpora in raw size, but to provide a scalable recipe for constructing large multi-view datasets from real VHR remote-sensing imagery. GeomSat contains \(8.96\)M observations grouped into \(1.83\)M co-located multi-view samples, with \(4.89\) views per group on average, and is automatically generated through an open processing pipeline built on open-source tools.

Remote-sensing datasets with physically grounded 3D supervision, in contrast, have so far operated at a substantially smaller scale. US3D covers approximately \(100\,\mathrm{km}^2\), TLC SatMVS contains 519 images, S-EO approximately \(20{,}000\) images, and GlobalBuildingAtlas \(187{,}239\) LiDAR-derived image--height pairs~\cite{bosch2019semantic,gao2021satmvs,masquil2025seo,zhu2025globalbuildingatlas}. In contrast, GeomSat provides spatially distributed multi-view observations worldwide and complements them with LiDAR-supervised views. From the aligned LiDAR and refined sensor geometry, we automatically derive dense height, surface-normal, and above-height displacement fields. In addition, these annotations are generated directly in the geometry of each off-nadir acquisition, rather than only as nadir-referenced DSM products, providing multiple complementary geometric supervision signals at scale. 
Table~\ref{tab:dataset_comparison} summarizes this comparison in terms of scale, views, and geometric supervision.


\begin{table*}[t]
    \centering
    \caption{\textbf{Positioning of GeomSat with respect to representative geometric learning datasets and methods.}
    ``Scale'' follows the unit reported by each work and is therefore not directly comparable across rows.
    The ``Views / sample'' column distinguishes single-image, pairwise, and multi-view training structures; GeomSat contains \(4.89\) aligned observations per multi-view group on average.
    Checkmarks indicate supervision or geometric information explicitly available during training.
    GeomSat combines large-scale multi-view pretraining with a LiDAR-supervised subset providing dense metric height, surface normals, and a two-dimensional above-height displacement field.}
    \label{tab:dataset_comparison}

    \setlength{\tabcolsep}{4.0pt}
    \renewcommand{\arraystretch}{1.15}

    \resizebox{\textwidth}{!}{
    \begin{tabular}{
        c
        l
        l
        c
        @{\hspace{3pt}}|
        @{\hspace{2pt}}c
        @{\hspace{2pt}}c
        @{\hspace{2pt}}c
        @{\hspace{2pt}}c
        @{\hspace{2pt}}c
        @{\hspace{2pt}}c
        @{\hspace{2pt}}c
    }
        \toprule
        &
        \textbf{Work} &
        \textbf{Scale} &
        \textbf{Views / sample} &
        \rotatebox{60}{\textbf{Cross-view}} &
        \rotatebox{60}{\textbf{Dense 3D}} &
        \rotatebox{60}{\textbf{Metric}} &
        \rotatebox{60}{\textbf{LiDAR}} &
        \rotatebox{60}{\textbf{Normals}} &
        \rotatebox{60}{\textbf{Camera}} &
        \rotatebox{60}{\textbf{Above-height disp.}} \\
        \midrule

        \multirow{6}{*}{\rotatebox{90}{\textbf{Generic}}}
        & CroCo-v2~\cite{weinzaepfel2023croco}
        & \(5.3\)M real pairs
        & 2
        & \checkmark
        & --
        & --
        & --
        & --
        & --
        & -- \\

        & DUSt3R~\cite{wang2024dust3r}
        & \(8.5\)M pairs
        & 2
        & \checkmark
        & \checkmark
        & --
        & --
        & --
        & \checkmark
        & -- \\

        & Metric3D-v2~\cite{hu2024metric3dv2}
        & \(>16\)M images
        & 1
        & --
        & \checkmark
        & \checkmark
        & mixed
        & \checkmark
        & \checkmark
        & -- \\

        & Depth Anything V2~\cite{yang2024depthanythingv2}
        & 595k synthetic + 62M pseudo-labeled real
        & 1
        & --
        & \checkmark
        & --
        & --
        & --
        & --
        & -- \\

        & VGGT~\cite{wang2025vggt}
        & Large multi-dataset mixture
        & \(2\)--\(24\)
        & \checkmark
        & \checkmark
        & --
        & mixed
        & --
        & \checkmark
        & -- \\

        & VGGT-\(\Omega\)~\cite{wang2026vggtomega}
        & \(15\times\) VGGT sup. + video
        & Multi
        & \checkmark
        & \checkmark
        & --
        & mixed
        & --
        & \checkmark
        & -- \\

        \midrule

        \multirow{6}{*}{\rotatebox{90}{\textbf{Remote sensing}}}
        & US3D~\cite{bosch2019semantic}
        & 69 WV3 acq., \(\sim100\,\mathrm{km}^2\)
        & Multi
        & \checkmark
        & \checkmark
        & \checkmark
        & \checkmark
        & --
        & \checkmark
        & -- \\

        & TLC SatMVS~\cite{gao2021satmvs}
        & 519 images / 173 scenes
        & 3
        & \checkmark
        & \checkmark
        & \checkmark
        & \checkmark
        & --
        & \checkmark
        & -- \\

        & S-EO~\cite{masquil2025seo}
        & \(\sim20\)k images / 702 tiles
        & Multi
        & \checkmark
        & \checkmark
        & \checkmark
        & \checkmark
        & --
        & \checkmark
        & -- \\

        & GlobalBuildingAtlas~\cite{zhu2025globalbuildingatlas}
        & 187k image--nDSM pairs
        & 1
        & --
        & \checkmark
        & \checkmark
        & \checkmark
        & --
        & --
        & -- \\[2pt]

        \cmidrule(lr){2-11}
        
        & \textbf{GeomSat}
        & \(\mathbf{1.83}\)M groups / \(\mathbf{8.96}\)M views
        & \(\mathbf{4.89}\) avg.
        & \checkmark
        & --
        & --
        & --
        & --
        & \checkmark
        & -- \\

        & \textbf{GeomSat (w/ LiDAR)}
        & \(\mathbf{\sim3}\)M total labeled views
        & \(\mathbf{4.89}\) avg.
        & \checkmark
        & \checkmark
        & \checkmark
        & \checkmark
        & \checkmark
        & \checkmark
        & \checkmark \\

        \bottomrule
    \end{tabular}}
\end{table*}

\section{Method and architectural details}

\pparagraph{Acquisition metadata.}
For metadata-conditioned height estimation and correspondence prediction, each image is represented by 11 scalar features: the sine and cosine of the image orientation relative to true north (geographic bearing), measured clockwise along the downward image axis (2), the across-track and along-track viewing angles (2), the off-nadir angle (1), the solar elevation (1), the sine and cosine of the solar azimuth expressed in image coordinates (2), the ground sampling distance (GSD, 1), and the sine and cosine of the applied image rotation (2). Geographic bearing is measured clockwise from true north, while image-coordinate solar azimuth is measured clockwise from the negative row axis. The viewing angles and off-nadir angle are divided by $30^\circ$, and solar elevation by $90^\circ$, while GSD is expressed directly in metres per pixel. Under augmentatiom, GSD is updated according to the applied crop scale, the acquisition viewing angles and solar elevation retain their physical values, while the rotation pair records the applied image rotation and equals $(0,1)$ without rotation. This encoding uses camera and acquisition information independently of the supervised height values. Each image's vector is supplied to the stage-specific FiLM modules described below.

\pparagraph{DPT head and metadata conditioning for monocular height estimation.}
We use a four-stage DPT-style decoder~\cite{ranftl2021vision} connected to encoder blocks 3, 6, 9, and 12. For a $512\times512$ input, each stage provides $1024$ patch tokens with 768 channels. A separate LayerNorm ($\epsilon=10^{-5}$) normalizes each stage, whose tokens are reshaped to a $32\times32$ feature map and projected via convolutions to 128, 256, 512, and 512 channels, respectively. When acquisition metadata is enabled, each projected map is modulated before spatial resampling using FiLM~\cite{perez2018film}. A stage-specific MLP, followed by a ReLu activation and a linear layer, maps the 11-dimensional metadata vector to channel-wise parameters $(\boldsymbol\gamma_s,\boldsymbol\beta_s)$. Its final layer is zero-initialized, making the initial modulation an identity mapping. The four maps are resampled to $128\times128$, $64\times64$, $32\times32$, and $16\times16$ using, respectively, through convolutional upsampling steps. Coarse-to-fine fusion starts at the coarsest level and progressively incorporates finer skip features. The DPT head contains no batch normalization, and the no-metadata ablation omits the FiLM modules.

The monocular height estimation DPT models predict $C_{\mathrm{out}}=1$, 3, 4, or 6 channels for height alone, height with displacement, height with normals, or all three targets, respectively. The normal-supervised model used in our main comparisons therefore produces four channels.

\begin{table*}[t]
    \centering
    \caption{\textbf{RSME (m, $\downarrow$) results for MHE on GeomSat and US3D/DFC19.}
    Scoresver all retained pixels (All) or those with GT height
    $\geq20$\,m, reported for the full set and separately for pixels without
    and with vegetation. Aligned predictions use a signed scale and offset fitted
    to test GT per image. Best and second-best values per column are
    \textbf{bold} and \underline{underlined}. RS3DAda was trained on crops
    from the same US3D products, sharing their geographic coverage, illumination,
    and radiometric conditions. $^\dagger$CHMv2 specializes in canopy height.}
    \label{tab:height-benchmark-rsme}

    \footnotesize
    \setlength{\tabcolsep}{2.5pt}
    \renewcommand{\arraystretch}{1.04}
    \begin{tabular*}{\textwidth}{
        @{\extracolsep{\fill}}lr*{6}{r}@{\hspace{8pt}}*{6}{r}@{}
    }
        \toprule
        & & \multicolumn{6}{c}{\textbf{GeomSat}}
          & \multicolumn{6}{c}{\textbf{US3D/DFC19}} \\
        \cmidrule(lr){3-8}\cmidrule(lr){9-14}
        & & \multicolumn{2}{c}{Full}
          & \multicolumn{2}{c}{No vegetation}
          & \multicolumn{2}{c}{Vegetation}
          & \multicolumn{2}{c}{Full}
          & \multicolumn{2}{c}{No vegetation}
          & \multicolumn{2}{c}{Vegetation} \\
        \cmidrule(lr){3-4}\cmidrule(lr){5-6}\cmidrule(lr){7-8}
        \cmidrule(lr){9-10}\cmidrule(lr){11-12}\cmidrule(lr){13-14}
        Model & Params (M)
        & All & $\geq20$ & All & $\geq20$ & All & $\geq20$
        & All & $\geq20$ & All & $\geq20$ & All & $\geq20$ \\
        \midrule
        \multicolumn{14}{@{}l}{\textit{Predictions aligned to GT per image}} \\
        Depth Anything V2-B & 97.5
        & 3.05 & 11.63 & 2.08 & 16.96 & 4.25 & 11.42
        & \textbf{3.85} & \underline{14.68} & \textbf{3.46} & 15.19 & 5.50 & 12.47 \\
        VGGT & 1256.5
        & 3.09 & 11.77 & 2.17 & 17.60 & 4.26 & 11.54
        & \underline{3.99} & 14.74 & \underline{3.62} & \underline{15.16} & 5.58 & 12.99 \\
        MoGe-3-L & 370.3
        & 3.11 & 12.00 & 2.17 & 17.77 & 4.30 & 11.77
        & 4.23 & 15.68 & 3.88 & 16.21 & 5.77 & 13.43 \\
        VGGT-$\Omega$ & 1143.8
        & 3.14 & 12.03 & 2.19 & 16.96 & 4.33 & 11.84
        & 4.09 & \textbf{14.56} & 3.69 & \textbf{14.82} & 5.79 & 13.47 \\
        Metric3D-v2-L & 411.9
        & 3.28 & 13.10 & 2.28 & 20.91 & 4.54 & 12.78
        & 5.64 & 25.37 & 5.47 & 27.21 & 6.48 & 16.41 \\
        Depth Anything 3-B & 135.4
        & 3.29 & 13.08 & 2.30 & 20.98 & 4.53 & 12.76
        & 5.34 & 22.98 & 5.11 & 24.39 & 6.45 & 16.40 \\
        RS3DAda & 366.5
        & 2.69 & 10.31 & 1.76 & \underline{11.09} & 3.81 & 10.28
        & -- & -- & -- & -- & -- & -- \\
        CHMv2$^\dagger$ & 336.9
        & \underline{2.28} & \underline{8.20} & 1.83 & 21.98
        & \underline{2.92} & \underline{7.36}
        & 5.45 & 26.53 & 5.60 & 29.25 & \textbf{4.61} & 10.92 \\
        \textbf{Ours} & 137.4
        & -- & -- & -- & -- & -- & --
        & 4.19 & 17.86 & 4.09 & 19.33 & \underline{4.67} & 10.34 \\
        \midrule
        \multicolumn{14}{@{}l}{\textit{Metric predictions}} \\
        CHMv2$^\dagger$ & 336.9
        & 2.64 & 9.32 & 2.04 & 25.27 & 3.46 & 8.34
        & 7.26 & 36.04 & 7.06 & 40.10 & 8.28 & \underline{10.28} \\
        RS3DAda & 366.5
        & 3.75 & 19.41 & \textbf{1.29} & 13.74 & 5.99 & 19.56
        & 4.45 & 19.08 & 4.10 & 19.81 & 6.00 & 15.92 \\
        \textbf{Ours} & 137.4
        & \textbf{2.20} & \textbf{4.62} & \underline{1.70}
        & \textbf{5.34} & \textbf{2.88} & \textbf{4.60}
        & 6.86 & 19.54 & 6.36 & 21.33 & 9.08 & \textbf{9.92} \\
        \bottomrule
    \end{tabular*}
    \par\smallskip
\end{table*}
\pparagraph{Lightweight convolutional head.}
For frozen-encoder transfer experiments, we attach a convolutional decoder to the final encoder feature map. We first project the backbone's embedding dimension to 256 channels, followed by batch normalization and ReLU. Four upsampling stages each apply a $2\times2$ transposed convolution with stride 2, preserving 256 channels, followed by batch normalization and ReLU. For our patch-16 encoder, these stages increase spatial resolution from $32\times32$ to $64\times64$, $128\times128$, $256\times256$, and $512\times512$. A final $1\times1$ convolution produces one height channel for MHE or two background/building logits for WHU segmentation. The head uses only final-layer spatial features, without intermediate encoder features or skip connections. The same decoder width and upsampling design are used across backbones, adapting the input projection to their embedding dimension and the output projection to the task. No acquisition metadata is provided in the frozen-encoder results reported in the main paper.

\pparagraph{Dense correspondence head.}
We replace XoFTR's convolutional backbone with a feature pyramid built from encoder blocks 4, 8, and 12. For a $512\times512$ input, each stage provides a $32\times32$ map with 768 channels, normalized by a stage-specific LayerNorm ($\epsilon=10^{-6}$). Transposed convolutions with kernel/stride 8, 4, and 2 project these maps to 128, 196, and 256 channels at $1/2$, $1/4$, and $1/8$ input resolution, respectively. Each pyramid level is conditioned on its image's 11-dimensional acquisition metadata using the identity-initialized FiLM design described above. The encoder, projections, and FiLM modules are shared between the two images. We retain XoFTR's coarse matching, local refinement, and subpixel prediction modules~\cite{xoftr}. The encoder is initialized from our cross-view pre-training checkpoint, and the entire matching model is jointly fine-tuned.

\section{Additional experiments}

\begin{figure*}[t]
    \centering
    \includegraphics[width=\textwidth]{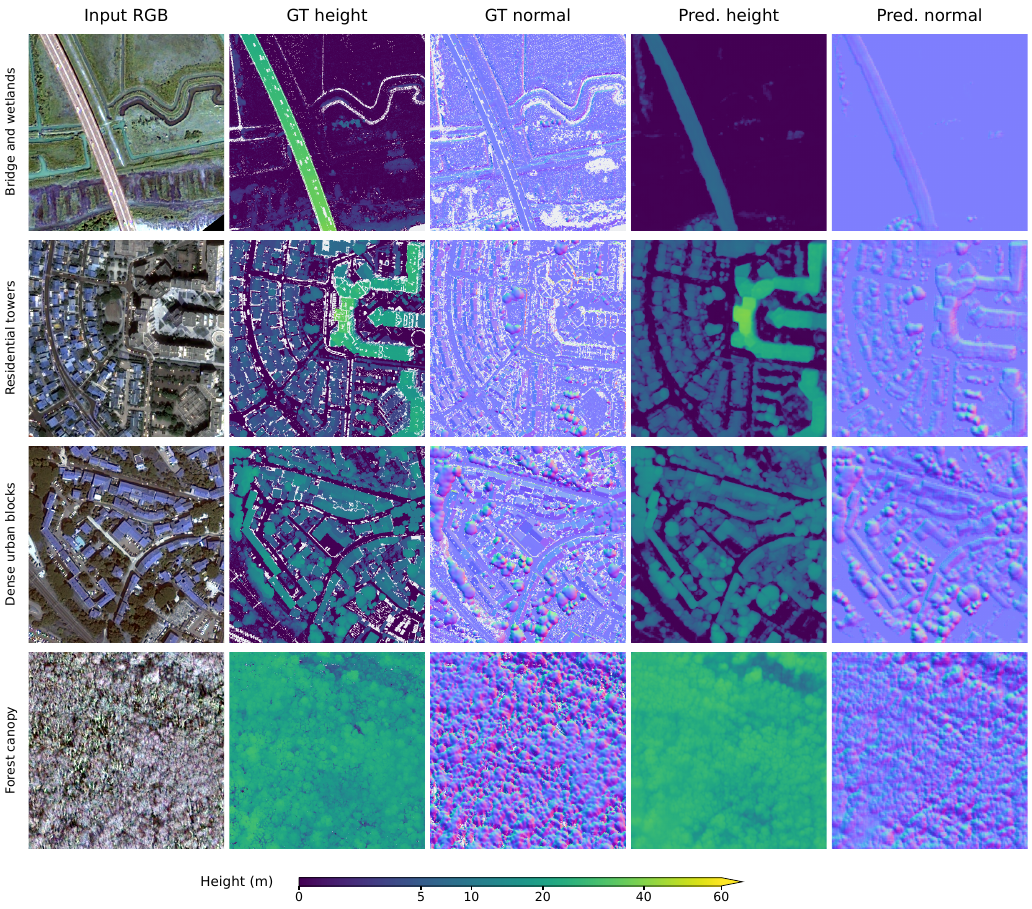}
    \caption{\textbf{Qualitative examples of the predictions of the proposed model on GeomSat examples}. For each input image, its ground truth height map and normal fields are shown, followed by the height and normal predictions of our model. The white pixels in the ground truths correspond to undefined values and thus are not used during training.}
    \label{fig:qual_france}
\end{figure*}

\begin{figure*}[t]
    \centering
    \includegraphics[width=\textwidth]{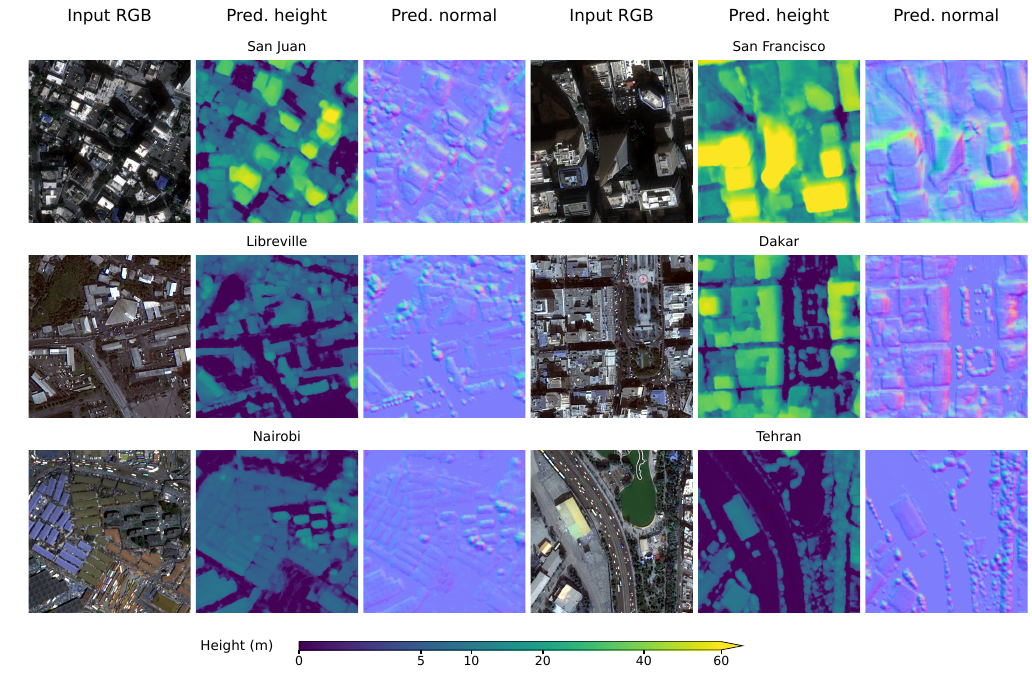}
    \caption{\textbf{Qualitative examples of the predictions of the proposed model on unseen geographical areas}. Input image and their corresponding height and normals predictions are shown for different areas of the globe. We have no geometric labels for these areas, as no LiDAR products associated to their localization are available. The san Francisco case is particularly challenging, as it contains uncommonly tall buildings with unconventional shapes.}
    \label{fig:qual_world}
\end{figure*}

\begin{figure*}[t]
    \centering
    \includegraphics[width=\textwidth]{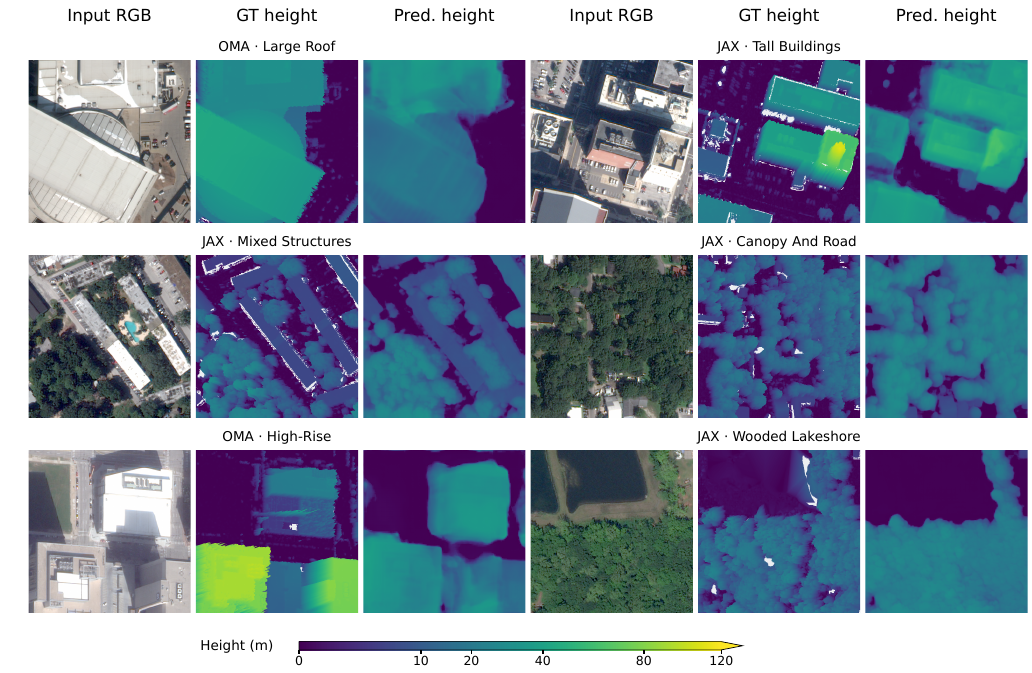}
    \caption{\textbf{Qualitative examples of the predictions of the proposed model on US3D/DFC19 examples (cross sensor scenario)}. For each input image, its ground truth height map and the height prediction of the proposed model is shown. OMA and JAX indicate Omaha and Jacksonville products, respectively.}
    \label{fig:qual_us3d}
\end{figure*}

\pparagraph{Qualitative examples.} We provide additional qualitative examples of the prediction of our monocular height estimation model on GeomSat images and US3D/DFC19. For the first, we show examples of images and the height and normals ground truth and prediction in Figure~\ref{fig:qual_france}, Additionally, Figure~\ref{fig:qual_world} shows prediction examples for unseen geographical areas for which we have no ground truth lidar data. Lastly, Figure~\ref{fig:qual_us3d} illustrates prediction examples in additional US3D/DFC19 samples.

\pparagraph{RSME general results.}
We provide in Table~\ref{tab:height-benchmark-rsme} the RSME results of the monocular height estimation (MHE) experiments for the GeomSat test set and US3D/DFC19.The results follow a similar direction as MAE scores, where we we generally outperform other methods on GeomSat, while cross sensor results are quite mixed. However, we score the best RMSE on tall vegetation areas for US3D/DFC19.

\pparagraph{Effect of pre-trained encoder for the MHE task.} We analyze the impact of the GeomSat pre-trained representations on the finetuning experiments for the monocular height estimation task. To this end, we compare the results of the MHE finetuning using three different encoder initializations: random weights, CroCo v2, and GeomSat. In all cases, we use the ViT-B/16 encoder architecture with the DPT head, trained on the 150k image set with height-only supervision and no acquisition metadata. As shown, our pre-trained representations show an advantage over the two other weight initializations, on both MAE and RMSE. This comparison with equal parameter counts illustrate the benefit of the GeomSat pre-trained representations in the full finetuning height estimation scenario.

\begin{table}[t]
    \centering
    \caption{\textbf{Effect of encoder initialization on MHE.}
    All models use ViT-B/16 with a DPT head, trained on 150k
    images with height-only supervision and no acquisition
    metadata. Errors are in metres ($\downarrow$).
    Each population is evaluated over all heights and over
    ground-truth heights $\geq20$\,m.
    Bold indicates the best available result.}
    \label{tab:mhe-encoder-initialization}
    \small
    \setlength{\tabcolsep}{2.5pt}
    \renewcommand{\arraystretch}{1.08}
    \begin{tabular}{@{}lrrrrrr@{}}
        \toprule
        & \multicolumn{2}{c}{All pixels}
        & \multicolumn{2}{c}{No veg.}
        & \multicolumn{2}{c}{Veg.} \\
        \cmidrule(lr){2-3}
        \cmidrule(lr){4-5}
        \cmidrule(lr){6-7}
        Initialization
        & All & $\geq20$ & All & $\geq20$ & All & $\geq20$ \\
        \midrule
        \multicolumn{7}{@{}l}{\textit{MAE} $\downarrow$} \\
        Random weights & 1.79 & 6.29 & 1.31 & 8.89 & 2.65 & 6.21 \\
        CroCo v2~\cite{weinzaepfel2023croco}
        & 1.23 & 5.14 & 0.69 & 6.21 & 2.19 & 5.11 \\
        GeomSat (Ours)
        & \textbf{0.93} & \textbf{3.35}
        & \textbf{0.49} & \textbf{3.92}
        & \textbf{1.70} & \textbf{3.33} \\
        \midrule
        \multicolumn{7}{@{}l}{\textit{RMSE} $\downarrow$} \\
        Random weights & 3.60 & 7.67 & 3.30 & 11.24 & 4.07 & 7.53 \\
        CroCo v2~\cite{weinzaepfel2023croco}
        & 2.86 & 6.45 & 2.43 & 8.41 & 3.48 & 6.37 \\
        GeomSat (Ours)
        & \textbf{2.24} & \textbf{4.70}
        & \textbf{1.79} & \textbf{5.99}
        & \textbf{2.86} & \textbf{4.65} \\
        \bottomrule
    \end{tabular}
\end{table}

\begin{figure*}[t]
    \centering
    \includegraphics[width=\textwidth]{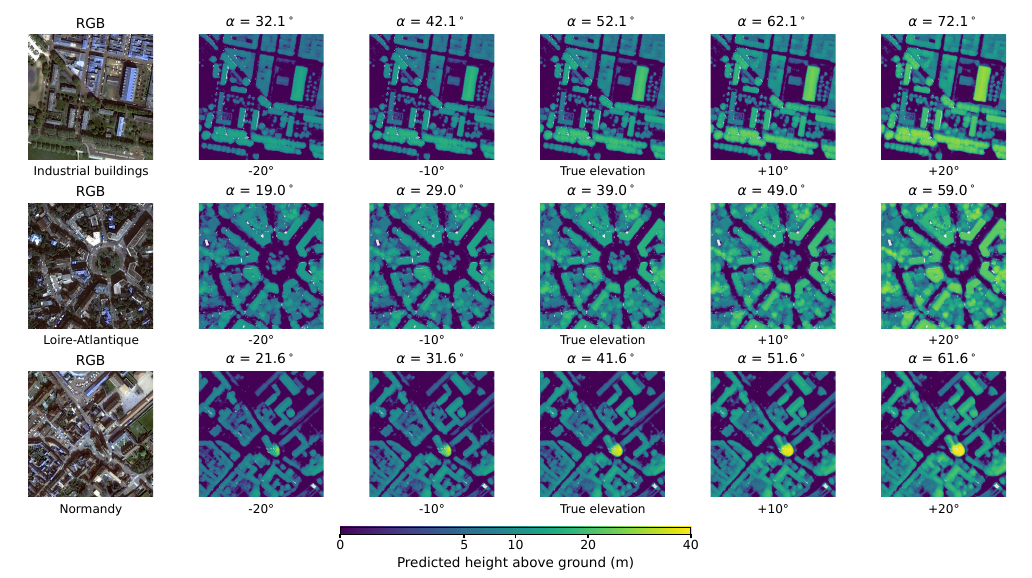}
    \caption{\textbf{Qualitative sensitivity to sun-elevation metadata.}
    We keep each image and all other metadata fixed while varying the supplied sun elevation. Column headings indicate offsets from the recorded elevation, and panel titles give the supplied absolute angle $\alpha$. All predictions share the same height scale. Increasing the sun elevation progressively increases predicted heights, while decreasing it produces lower estimates.}
    \label{fig:sun_elevation_maps}
\end{figure*}

\begin{figure*}[t]
    \centering
    \includegraphics[width=\textwidth]{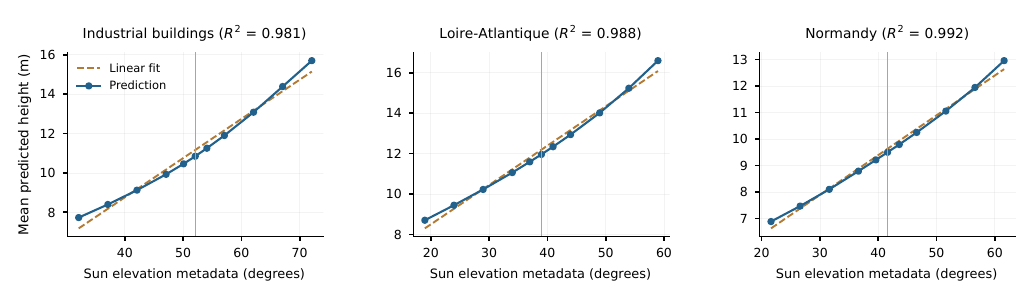}
    \caption{\textbf{Height response to sun-elevation metadata.}
    Blue points show mean predicted height over a fixed set of LiDAR-valid pixels with reference height $\geq5$\,m. Dashed lines show linear fits, and vertical grey lines mark the recorded sun elevations. The reported $R^2$ values measure the linearity of the response.}
    \label{fig:sun_elevation_response}
\end{figure*}

\pparagraph{Did the model learn to use metadata values?}
Beyond the metadata ablation in the main text, we provide a qualitative experiment to examine whether the model has learned to exploit acquisition metadata for height estimation. Shadows cast by tall structures provide an important cue, but interpreting their length requires knowledge of the sun elevation: for a fixed shadow length, a higher sun elevation implies a taller structure. In Figure~\ref{fig:sun_elevation_maps}, we examine whether the model responds consistently with this relationship by varying the supplied sun elevation while keeping the image and all other metadata fixed.

We use three test GeomSat Pleiades images from different regions and acquisition products. For each image, we evaluate thirteen sun elevations spanning $\pm30^\circ$ around the recorded value. Figure~\ref{fig:sun_elevation_maps} shows seven of these settings, using the same height scale across all predictions. Increasing the supplied sun elevation progressively increases the predicted heights, while decreasing it produces lower estimates. We quantify this behaviour in Figure~\ref{fig:sun_elevation_response} by measuring the mean predicted height over a fixed set of LiDAR-valid pixels with reference height above 5\,m. The response is monotonic and approximately linear in all three examples, with $R^2$ values between 0.977 and 0.982. These results indicate that the model has learned to use sun-elevation metadata in a direction consistent with shadow-based height geometry. The physical relationship is $h=L\tan\alpha$ for a fixed horizontal shadow length $L$ on level ground, so the observed linearity in elevation should be interpreted as an approximate response rather than evidence of exact geometric calibration.

\pparagraph{Additional ablations.}
We provide in Table~\ref{tab:height-ablation-vegetation2} complementary ablations that helped choose certain hyperparameters in our work. Panel (a) illustrates the improvement achieved with metadata and the specified image augmentations using the lightweight convolutional decoder. Similarly to the DPT head, adding metadata information and image augmentations improves substantially the monocular height estimation results.

In the panel (b), we compare the results under different peak learning rate values and combinations for the pre-trained encoder and head. We found the best combination to be a  $10^{-5}$ / $10^{-4}$ peak learning rates for the encoder and head, respectively. This achieved substantial improvement on non-vegetation areas for tall pixels, while the rest of cases remained quite similar in terms of MAE.

\begin{figure}[t]
    \centering
    \includegraphics[width=\columnwidth]{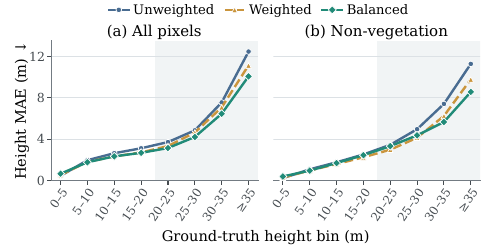}
    \caption{\textbf{Effect of height-bin balancing.}
    GeomSat MAE per ground-truth height bin for all pixels (left)
    and non-vegetation pixels (right).
    Balancing reduces errors on elevated structures with a trade-off in the lowest bin.
    Shading indicates heights $\geq20$\,m.}
    \label{fig:height-bin-balancing}
\end{figure}

Panel (c) illustrates the effect of weighting pixels by their ground-truth height in 5,m bins to reduce the dominance of the 0--5,m range. HM + weighting applies a mild correction, whereas HM + balanced weighting, used in our main experiments, assigns inverse-frequency weights based on training-set pixel counts, giving each bin equal aggregate weight in the loss. This balancing improves height estimation for tall structures at the cost of slightly higher errors in low-height regions. Figure~\ref{fig:height-bin-balancing} shows this trade-off through test-set MAE for each height bin: errors decrease substantially in the higher bins, with a modest increase in the 0--5,m bin.

Lastly, panel (d) of Table~\ref{tab:height-ablation-vegetation2} shows the improvement of the DPT head over the lightweight convolutional head. In this case, we add to the convolutional head one FiLM layer to condition the model on the metadata, similarly to the DPT case. 
Replacing the convolutional head with DPT reduces overall MAE from 0.99 to 0.93\,m and MAE above 20\,m from 3.09 to 2.92\,m. The improvement holds across vegetation and non-vegetation: their overall MAEs decrease from 1.76 to 1.69\,m and from 0.56 to 0.50\,m, respectively. These results support using multi-scale encoder features through the DPT head for dense height prediction.

\begin{table*}[t]
    \centering
    \caption{\textbf{Ablations for augmentations, learning rates, height-loss
    weighting, and head architecture.}
    Test MAE in meters ($\downarrow$) on the GeomSat test set.
    MAE$_{\geq20}$ pools pixels with ground-truth height $\geq20$\,m.
    Crop/color augmentation uses random resized crops, random rotations in $[-180^\circ,180^\circ]$ and RGB gain/contrast.
    Blocks (a) and (b) use the lightweight convolutional head.
    Block (d) uses inverse-frequency height-bin balancing,
    crop/color/rotation augmentation, and peak encoder/head learning
    rates $10^{-5}/10^{-4}$.}
    \label{tab:height-ablation-vegetation2}
    \setlength{\tabcolsep}{5pt}
    \renewcommand{\arraystretch}{1.08}
    \small
    \begin{tabular}{@{}lrrrrrr@{}}
        \toprule
        & \multicolumn{2}{c}{Full}
        & \multicolumn{2}{c}{No vegetation}
        & \multicolumn{2}{c}{Vegetation only} \\
        \cmidrule(lr){2-3}
        \cmidrule(lr){4-5}
        \cmidrule(lr){6-7}
        Model
        & MAE & MAE$_{\geq20}$
        & MAE & MAE$_{\geq20}$
        & MAE & MAE$_{\geq20}$ \\
        \midrule
        \multicolumn{7}{@{}l}{\textit{(a) Metadata and augmentation}} \\
        \addlinespace[2pt]
        Height
            & 0.77 & 4.15
            & \textbf{0.27} & 4.31
            & 1.66 & 4.14 \\
        Height + metadata (HM)
            & 0.76 & 3.97
            & 0.28 & 4.39
            & 1.62 & 3.96 \\
        HM + augmentations
            & 0.76 & \textbf{3.87}
            & \textbf{0.27} & 4.22
            & 1.62 & \textbf{3.86} \\
        \midrule
        \multicolumn{7}{@{}l}{\textit{(b) Peak learning rates
        (encoder / head); HM + crop/color}} \\
        \addlinespace[2pt]
        $10^{-4}$ / $10^{-4}$
            & \textbf{0.76} & \textbf{3.87}
            & \textbf{0.27} & 4.22
            & 1.62 & \textbf{3.86} \\
        $3\!\times\!10^{-5}$ / $3\!\times\!10^{-5}$
            & \textbf{0.76} & 4.15
            & \textbf{0.27} & 4.12
            & 1.62 & 4.15 \\
        $10^{-5}$ / $10^{-5}$
            & 0.77 & 4.08
            & \textbf{0.27} & 4.03
            & 1.64 & 4.08 \\
        $10^{-5}$ / $10^{-4}$
            & \textbf{0.76} & 3.99
            & 0.28 & \textbf{3.84}
            & \textbf{1.60} & 3.99 \\
        \midrule
        \multicolumn{7}{@{}l}{\textit{(c) Height-loss weighting;
        HM without augmentation}} \\
        \addlinespace[2pt]
        HM, unweighted
            & \textbf{0.76} & 3.97
            & \textbf{0.28} & 4.39
            & \textbf{1.62} & 3.96 \\
        HM + weighting
            & 0.79 & 3.61
            & 0.32 & \textbf{3.76}
            & \textbf{1.62} & 3.60 \\
        HM + balanced weighting
            & 0.90 & \textbf{3.37}
            & 0.43 & 3.94
            & 1.72 & \textbf{3.36} \\
        \midrule
        \multicolumn{7}{@{}l}{\textit{(d) Head architecture;
        balanced height loss, with augmentations}} \\
        \addlinespace[2pt]
        HM, Conv. head
            & 0.99 & 3.09
            & 0.56 & 3.59
            & 1.76 & 3.08 \\
        HM, DPT head
            & \textbf{0.93} & \textbf{2.92}
            & 0.50 & \textbf{3.35}
            & \textbf{1.69} & \textbf{2.90} \\
            \bottomrule
    \end{tabular}
\end{table*}

\end{document}